\documentclass[10pt,twocolumn,letterpaper]{article}

\usepackage[pagenumbers]{cvpr} 

\usepackage{xcolor,colortbl}

\usepackage{pgfplots, tikz}
\usepackage{blindtext,tikz}
\usepackage{booktabs}
\usepackage{multirow}
\usepackage{graphicx}

\usepackage{pgfplotstable}
\usetikzlibrary{patterns}
\usepgfplotslibrary{fillbetween}

\definecolor{cb-blue}{HTML}{0072B2}
\definecolor{cb-orange}{HTML}{E69F00}
\definecolor{cb-green}{HTML}{009E73}
\definecolor{cb-red}{HTML}{D55E00}

\definecolor{cb-one}{HTML}{648FFF}
\definecolor{cb-two}{HTML}{DC3220}
\definecolor{cb-three}{HTML}{40B0A6}
\definecolor{cb-four}{HTML}{FFB000}

\definecolor{lightgray}{gray}{0.9}

\definecolor{cvprblue}{rgb}{0.21,0.49,0.74}
\usepackage[pagebackref,breaklinks,colorlinks,allcolors=cvprblue]{hyperref}

\def\paperID{42} 
\def\confName{CVPR}
\def\confYear{2025}

\title{Learning from Noise: Enhancing DNNs for Event-Based Vision through Controlled Noise Injection}

\author{Marcin Kowalczyk, Kamil Jeziorek, Tomasz Kryjak\\
Embedded Vision Systems Group, AGH University of Krakow, Poland\\
{\tt\small kowalczyk@agh.edu.pl kjeziorek@agh.edu.pl tomasz.kryjak@agh.edu.pl}
}

\usetikzlibrary{calc}

\begin{document}

\maketitle

\tikz[overlay,remember picture] {
    \node at ($(current page.north)+(0,-1.5)$) {\textcolor{gray}{This paper has been accepted for publication at the}};
    \node at ($(current page.north)+(0,-2.0)$) {\textcolor{gray}{IEEE Conference on Computer Vision and Pattern Recognition (CVPR) Workshops, Nashville, 2025. ©IEEE}};
}

\begin{abstract}

Event-based sensors offer significant advantages over traditional frame-based cameras, especially in scenarios involving rapid motion or challenging lighting conditions. However, event data frequently suffers from considerable noise, negatively impacting the performance and robustness of deep learning models.
Traditionally, this problem has been addressed by applying filtering algorithms to the event stream, but this may also remove some of relevant data.
In this paper, we propose a novel noise-injection training methodology designed to enhance the neural networks robustness against varying levels of event noise.
Our approach introduces controlled noise directly into the training data, enabling models to learn noise-resilient representations.
We have conducted extensive evaluations of the proposed method using multiple benchmark datasets (N-Caltech101, N-Cars, and Mini N-ImageNet) and various network architectures, including Convolutional Neural Networks, Vision Transformers, Spiking Neural Networks, and Graph Convolutional Networks. Experimental results show that our noise-injection training strategy achieves stable performance over a range of noise intensities, consistently outperforms event-filtering techniques, and achieves the highest average classification accuracy, making it a viable alternative to traditional event-data filtering methods in an object classification system.

Code: \href{https://github.com/vision-agh/DVS_Filtering}{https://github.com/vision-agh/DVS\_Filtering}
\end{abstract}    
\section{Introduction}
\label{sec:introduction}

Event cameras, also known as neuromorphic cameras, are advanced vision sensors offering high temporal resolution, low latency, and energy efficiency, making them a compelling alternative to conventional frame-based cameras \cite{4444573, 9063149}. They feature a unique operational mechanism that registers intensity changes independently for each pixel. This allows efficient acquisition of visual information in dynamic scenarios, such as robotics \cite{10209039}, autonomous vehicles \cite{Gehrig24nature}, or real-time tracking of fast-moving objects \cite{8593805}.
However, raw event data is often vulnerable to noise and artifacts resulting from environmental disturbances and hardware imperfections. Furthermore, the noise intensity varies according to the operating conditions of the sensor, including the brightness of the observed scene, the sensor temperature, and operational parameters. Consequently, event data processing algorithms should, to a certain extent, be resilient to fluctuations in noise levels.
It is also noteworthy that neural networks are increasingly employed for event data processing tasks, including image reconstruction, object detection, and classification. Among the methods frequently used are convolutional neural networks (CNNs) \cite{gehrig2019end, baldwin2019inceptive, messikommer2020event}, spiking neural networks (SNNs) \cite{dampfhoffer2024neuromorphic, chen2023sign, yao2021temporal, stromatias2017event, barchid2023spiking}, vision transformers (ViTs) \cite{sabater2022event, xie2024event, jiang2024token, de2023eventtransact}, and increasingly, graph convolutional networks (GCNs) \cite{schaefer2022aegnn, li2021graph, chen2020dynamic, bi2019graph, deng2022voxel}. It is evident that the presence of noise can negatively affect their performance. This effect can be mitigated by filtering algorithms \cite{delbruck2008frame, xu2023denoising, kowalczyk2023interpolation, baldwin2020event, guo2022low, Rios-Navarro_2023_CVPR, padala2018noise, duan2024led}. However, it should be noted that each filtering algorithm inevitably removes at least a small portion of true events, which may in turn reduce the effectiveness of the processing algorithms.

\begin{figure}[t]
    \centering
    \includegraphics[width=0.49\textwidth]{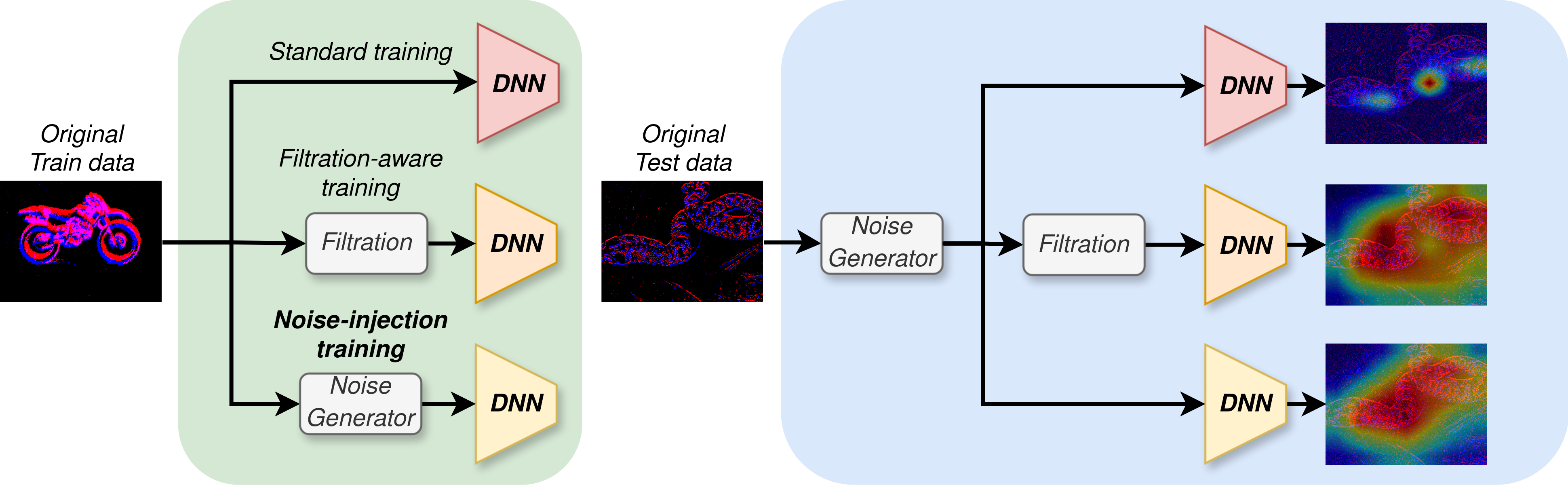}
    \caption{A simplified diagram of the proposed solution. Standard training without filtering leads to significant instability, while applying filters can lead to sensitivities due to excessive data reduction or retention of noise artefacts. Our noise-injection training method enhances generalisation by effectively managing noisy input data, thereby improving the robustness of neural networks in real-world conditions.} 
    \label{fig:teaser}
\end{figure}


The objective of this work is to investigate the impact of varying noise intensities on the performance of multiple types of deep neural network architectures in processing event-camera data. Additionally, we seek to verify the hypothesis that \textit{training neural networks on input data with different levels of noise intensity can improve their effectiveness under variable noise conditions}. Specifically, the employment of an approach that involves the exposure of the models to data of varying quality (demonstrated in \cref{fig:teaser}) may result in enhanced generalisation and increased robustness of the models against real-world environmental disturbances.

However, it is important to acknowledge that when a neural network is trained on data with different noise levels, part of its resources may be devoted to denoising tasks. This can potentially limit its maximum performance under ideal conditions, which is also a crucial aspect that requires further analysis and will be investigated in this study.

The main contribution of this paper can be summarised as follows:
\begin{itemize}
\item We propose a novel noise-injection event data augmentation method for neural networks by introducing shot noise at different intensities, aiming to reduce the impact of noise on network performance.
\item We analysed the effect of noise at different intensity levels on the performance of several popular neural network architectures (CNN, SNN, ViT, GCN) in a classification task, as well as the improvement achieved after applying our proposed augmentation technique.
\item We investigated the influence of three state-of-the-art event data filtering methods on the classification accuracy of various neural network architectures and evaluated their impact when combined with our proposed training methodology.
\end{itemize}

The rest of the article is organized as follows.
Section \ref{sec:related} analyses articles related to noise filtration, object classification on event-based data and event-data augmentation methods.
Section \ref{sec:method} describes the operation of the methods used in the paper.
Section \ref{sec:experiments} presents the experiments performed and the results obtained.
Section \ref{sec:discusion} discusses the results obtained throughout the research.
Finally, section \ref{sec:conclusion} summarises the main findings, draws conclusions, and outlines future directions.
\section{Related Work}
\label{sec:related}


Issues related to noise in event-based data and object classification have been frequently addressed in the scientific literature due to their significant practical importance. This section provides an overview of existing research relevant to the topic under consideration.

\subsection{Noise filtration}

A thorough analysis of pixel behaviour in event-based cameras was presented in \cite{gracca2023shining}. The authors also analysed how background activity intensity varies depending on the sensors operational parameters (bias settings).


The most popular algorithm for filtering event data to remove uncorrelated noise was proposed in \cite{delbruck2008frame}. This method involves the elimination of events lacking activity recorded in their immediate vicinity within a specified time window. This approach commonly known as the nearest-neighbour (NN) method. Similar solutions based on correlation and analysis of the event neighbourhood have also been presented in works such as \cite{liu2015design, czech2016evaluating, ojeda2020device, xu2023denoising, kowalczyk2023interpolation}.


Another approach involves employing neural networks to determine whether a given event is noise or a valid signal. The most popular algorithm of this type is the Event Denoising Convolutional Neural Network (EDnCNN) introduced in \cite{baldwin2020event}, which employs a CNN model comprising three convolutional and two fully connected layers. The aforementioned work also proposed a mechanism for assessing the probability of event generation by individual pixels, effectively enabling estimation of the likelihood that a given event is noise. Other methods based on classical neural networks include \cite{guo2022low, Rios-Navarro_2023_CVPR, fang2022aednet}.


The utilisation of spiking neural networks, which are well-suited for processing data from neuromorphic cameras, has also gained popularity for noise reduction. For instance, in \cite{padala2018noise}, the proposed network was deployed on the neuromorphic processor \textit{IBM TrueNorth}. This network consisted of two layers: the first introducing a refractory period and the second verifying whether additional events were generated near the processed event. Other approaches based on spiking neural networks were presented in \cite{xiao2021snn, duan2024led}.

\subsection{Classification}


Classification is a popular task in both classical image processing and event-based vision. Today, nearly all algorithms of this type rely on neural networks. CNNs are among the most common solutions \cite{gehrig2019end, baldwin2019inceptive, messikommer2020event}, leveraging convolutional layers to extract local features such as edges, textures, and more complex patterns. A network of this type was utilised in \cite{gehrig2019end}, where the authors proposed a framework to convert asynchronous event data to grid-based representations through a sequence of differentiable operations. In \cite{messikommer2020event}, synchronous convolutions were initially trained on image-like event representations (e.g. \textit{Event Count Image} \cite{maqueda2018event}) and subsequently adapted to asynchronous versions while preserving identical outputs.


Numerous solutions employing Vision Transformers have also been proposed for event-based classification tasks \cite{xie2024event, sabater2022event, jiang2024token, de2023eventtransact}. ViTs effectively capture global context by modelling image data as sequences of patches processed through a self-attention mechanism. This allows ViTs to model relationships even between distant regions of the image, thereby enhancing their ability to capture global patterns and contextual information. An example is the Event Transformer (EvT) proposed in \cite{sabater2022event}, where the authors introduced a transformer-based architecture utilising a novel patch-based representation that includes only pixels activated by events. Another example is the three-way attention mechanism \cite{jiang2024token} designed to preserve the spatio-temporal attributes of event data.


SNNs have also gained popularity in this task \cite{dampfhoffer2024neuromorphic, chen2023sign, yao2021temporal, stromatias2017event, barchid2023spiking}.
In \cite{dampfhoffer2024neuromorphic}, the authors employed an SNN for word classification based on lip movements, introducing a novel Signed Spiking Gated Recurrent Unit (SpikGRU2+) neuron as the task head. Authors of \cite{chen2023sign} explored the use of SNNs for sign language classification based on event camera data.


Another promising approach involves Graph Convolutional Networks, which have been designed for handling irregular and sparse data \cite{schaefer2022aegnn, li2021graph, chen2020dynamic, bi2019graph, deng2022voxel}. These networks effectively process event data modelled as evolving spatio-temporal graphs by employing graph convolutional operations. The authors of \cite{schaefer2022aegnn} proposed the AEGNN network, which formulates efficient update rules for newly arriving events. In contrast, \cite{deng2022voxel} employs a voxelisation strategy, preprocessing events by selecting representative voxels prior to the application of graph convolutions, thereby reducing computational complexity.


To the best of the authors knowledge, no previous work has systematically examined the impact of varying levels of noise on event-based neural network models. Moreover, there is no prior research that explicitly explores the influence of noise exposure during training on model robustness.

\subsection{Event Data Augmentation}


Data augmentation is a widely utilised technique that aims to enhance the generalisation and performance of machine learning models. Typical augmentation methods, well-established in computer vision, include basic geometric transformations such as translation, rotation, scaling, cropping, and colour space modifications \cite{NIPS2012_c399862d, perez2017effectiveness, yang2022image}. More advanced techniques involve combining different images into single training samples, as presented in \cite{zhang2017mixup, inoue2018data}.


In the context of event-based camera data, traditional augmentation techniques (e.g. translations, scaling, rotations) can be relatively easily adapted by generating new event streams through appropriate geometric transformations \cite{li2022neuromorphic}. More specific event-data augmentation techniques have also been introduced, such as EventMix \cite{SHEN2023119170}, which combines different event streams into single training samples, or EventDrop \cite{gu2021eventdrop}, a random event-removal method aimed at enhancing model generalisation by intentionally omitting parts of the event stream. Another notable technique described in the literature is EventRPG \cite{sun2024eventrpg}, which presents RPGDrop and RPGMix augmentations by dropping and mixing events with the guidance of saliency information of SNNs. 

However, to the best of the authors knowledge, no previous studies have investigated the direct application of variable shot noise levels as a data augmentation strategy specifically for event-based sensor data, nor evaluated its impact on different neural network models performance.


\section{Methods}
\label{sec:method}




This section provides an overview of the methods and tools utilised in the research, including the operating principle of the event camera, the noise modelling process, event filtering, the strategy for incorporating noise during training, a detailed description of the used datasets, the architectures of the considered neural networks, and their implementations.

\subsection{Event data and noise generation}


Each pixel in an event camera operates independently and responds to changes in light intensity. When the change in logarithmic light intensity at a given pixel exceeds a predefined threshold, a single event is generated: \(e_i = (x_i, y_i, t_i, p_i)\). The event is characterised by a timestamp \( t_i \), the pixel coordinates \((x_i, y_i)\), and a polarity \( p_i \), which can be either positive or negative depending on the direction of the intensity change. As a result, the event camera produces a sparse, spatio-temporal stream of events with high temporal resolution, triggered by changes occurring in the observed scene. 


However, these sensors are not perfect, and their outputs often contain noise artifacts. In the context of event-based sensors, shot noise arises from fluctuations in the pixel photoreceptor and change-detection circuits, which randomly cross the event-triggering threshold. This type of noise can be effectively modelled as a Poisson process. Such modelling was utilised in \cite{hu2021v2e} to generate noise during the conversion of frame-based data into event-based data. A similar approach was presented in \cite{guo2022low} for event data filtering, allowing for the evaluation of filtering performance.

Another method for introducing noise to event data involves the use of recordings obtained from a real-event sensor observing a static scene with constant illumination. In this scenario, all recorded events can be regarded as noise. These noise samples can then be added to other event sequences, thereby enabling the adjustment of noise levels. However, this approach is characterised by limited control over the precise noise intensity and may be influenced by micro-movements or slight illumination fluctuations present during the recording process.

In this work, we employ an approximation of the Poisson process with a sequence of Bernoulli trials to generate datasets with different noise levels. We assume that all pixels share the same average noise rate. The time step for this process is calculated based on the image resolution and the desired noise intensity as in the \cref{eq:Delta_t}.

\begin{equation}
    \Delta t = \frac{1}{\lambda \cdot N \cdot D}
\label{eq:Delta_t}
\end{equation}
where \(\lambda\) is the noise intensity [Hz/px], \(N\) is the number of pixels, and \(D\) is the parameter for dividing the time step length, allowing for a better approximation of the Poisson process.
The Poisson process is well approximated if the \cref{eq:PoissonCondition} is satisfied.

\begin{equation}
    P = \lambda \cdot \Delta t = \frac{1}{N \cdot D} \ll 1
\label{eq:PoissonCondition}
\end{equation}
where \(P\) is the probability of event generation for a pixel in each step.
A pseudo-random number determines whether an event should be generated in each \(\Delta t\), and if so, the pixel coordinates to which it is assigned are drawn randomly.

\subsection{Event noise filtering}

Event data filtering is a process that enables the elimination of events that are unrelated to genuine brightness variations in the observed scene. In this study, three filtering algorithms were applied to evaluate their impact on the classification performance of neural networks. The test data, augmented with synthetic noise, was filtered using three methods: the nearest neighbour (NN) \cite{delbruck2008frame}, the Event Denoising CNN (EDnCNN) \cite{baldwin2020event}, and the Distance-based Interpolation with Frequency weights (DIF) \cite{kowalczyk2023interpolation}.

The NN filter removes events that do not have neighbouring events within a \(3 \times 3\) window and a specified temporal window. The EDnCNN method uses a convolutional neural network, where for each event, a \(25 \times 25\) spatial neighbourhood is considered. For each pixel in this neighbourhood, the network processes information about the time elapsed since the last two positive and the last two negative events.  

The DIF algorithm, designed for low-memory embedded platforms, divides the sensor array into subregions, within which timestamps and intervals between consecutive events are filtered. The filtered timestamps are then interpolated based on the distance of the processed event to the centres of neighbouring subregions and their respective intervals. The difference between the event’s timestamp and the interpolated value determines whether the event is removed.  

The filtered test datasets were subsequently evaluated using trained models, and the classification performance was compared to results obtained without filtering.

\subsection{Noise-injection data augmentation}


The impact of training neural networks on noisy data was also investigated. Two approaches were employed.
In the first one, all training samples were augmented with noise of a fixed intensity. The networks trained in this manner were then evaluated on test datasets containing noise of varying intensities, both with and without prior filtering.

The second approach involved training on data with varying noise levels, using the same neural network architectures as in previous experiments. In this case, noise injection was treated as a form of data augmentation. For each loaded training sample, a random value was drawn to determine the corresponding noise intensity, including the possibility of no noise. The networks trained in this way were evaluated on test datasets with different noise levels, as well as on datasets that had been filtered beforehand.

\subsection{Model Architectures and Event Representations}

\begin{table}[tp]
\centering
\resizebox{0.7\columnwidth}{!}{%
\begin{tabular}{@{}ccc@{}}
\toprule
Model type           & Parameter name      & Parameter               \\ \midrule
\multirow{3}{*}{CNN} & Architecture        & ResNet18 \cite{he2016deep}               \\
                     & Input channels      & 2                       \\
                     & Backbone parameters & 11.17 M                 \\ \midrule
\multirow{5}{*}{ViT} & Architecture        & MaxViT \cite{tu2022maxvit}                 \\
                     & Input channels      & 20 \\ 
                     & Stages              & 4                       \\
                     & Depth of stages     & (1, 1, 1, 1)            \\
                     & Backbone parameters & 13.37 M                 \\ \midrule
\multirow{4}{*}{SNN} & Architecture        & ResNet18 \cite{he2016deep, spikingjelly}               \\
                     & Input channels      & 2                       \\
                     & Backbone parameters & 11.17 M                 \\ \midrule
\multirow{4}{*}{GCN} & Architecture        & GCN ResNet \cite{schaefer2022aegnn} \\
                     & Convolution         & SplineConv              \\
                     & Input Channels      & 1                       \\
                     & Backbone parameters & 8.52 M                  \\ \bottomrule
\end{tabular}%
}
\caption{Details of the evaluated models, including architecture types and the number of backbone parameters.}
\label{table:models}
\end{table}


To evaluate the impact of the noise and the effectiveness of the proposed noise-injection training approach, we conducted experiments using four widely recognised backbone architectures commonly applied in event-based data processing. \cref{table:models} summarises the detailed specifications of these models, including their architectures, input configurations, and the number of backbone parameters.


The first architecture is a Convolutional Neural Network (CNN) based on a modified ResNet18 \cite{he2016deep}, where the first layer was adapted to accommodate event-based data representations. We employed an \textit{Event Count Image} \cite{maqueda2018event} with two channels, each summing the number of events per polarity to create a representation of dimension \(2 \times W \times H\).


The second architecture, inspired by the work \cite{Gehrig_2023_CVPR}, involves a Vision Transformer (ViT) utilising MaxViT \cite{tu2022maxvit}, which incorporates MBConv blocks, block attention, and grid attention mechanisms per stage. Here, the \emph{Voxel Grid} approach was adopted following the same work: the event window is divided into T temporal intervals, each processed into separate \emph{Event Count Images}, combined into a representation with dimensions \(2T \times W \times H\).


Third model is a Spiking Neural Network (SNN), similarly employing a modified ResNet18 structure with \textit{Integrated Fire} neurons and the \textit{ATan} surrogate function. For SNN, we utilised the \emph{Event Spike Tensor} \cite{gehrig2019end}, dividing the event window into T intervals and summing event polarities, resulting in a tensor of dimensions \(T \times 2 \times W \times H\).


The final architecture explored is a~Graph Convolutional Network (GCN), based on the approach proposed by \cite{schaefer2022aegnn}, employing a residual structure with SplineConv layers \cite{fey2018splinecnn}. Due to substantial memory demands and computational complexity associated with increasing event numbers (noise levels), we adopted a spatio-temporal voxelisation strategy, generating a \emph{Voxel Graph} representation instead of creating a graph node per event. This strategy involves dividing the space into small voxels in which events are grouped into a single representative vertex with an averaged spatio-temporal position and average polarity.


Each model was complemented by a classifier consisting of a single fully connected layer, whose output dimensionality corresponds directly to the number of classes in the dataset considered.

\subsection{Datasets}



To evaluate our method, we used three popular datasets commonly employed for classification tasks: N-Caltech101 \cite{orchard2015converting}, N-Cars \cite{sironi2018hats}, and Mini N-ImageNet \cite{Kim_2021_ICCV}. The N-Caltech101 and Mini N-ImageNet datasets are event-based counterparts of the well-known frame-based Caltech101 \cite{1384978} and Mini ImageNet \cite{5206848, vinyals2016matching}, created by capturing images displayed on a monitor using event-based cameras. In contrast, the N-Cars dataset consists of real-world event data and is comparatively simpler, containing only two classes with a resolution of \(120 \times 100\). The N-Caltech101 and Mini N-ImageNet datasets each contain a larger number of classes and higher resolutions: N-Caltech101 includes 101 categories at a resolution of \(240 \times 180\), while Mini N-ImageNet has 100 categories with a resolution of \(640 \times 480\).

\input{sec/plots_training}

\subsection{Implementation Details}




All experiments were conducted using the PyTorch environment \cite{paszke2017automatic} with PyTorch Geometric \cite{Fey/Lenssen/2019} and SpikingJelly \cite{spikingjelly} libraries for the implementation of GCN and SNN models. In ViT and SNN models, the parameter \(T\) (time-steps) was set to 10. For the N-Caltech101 and Mini N-ImageNet, 50-ms event windows were adopted, while for the N-Cars, a 100-ms interval was used. In ViT model representations, images were resized to \(128 \times 128\) for N-Cars and \(224 \times 224\) for N-Caltech101 and Mini N-ImageNet.

To generate the graph, spatio-temporal positions were normalised to the range [0,1], with a search radius of 0.02 and a maximum number of neighbours set to 16. Before generating representations, the events were transformed with Random HorizontalFlip, Crop, and Translate augmentations. The AdamW optimiser \cite{loshchilov2017decoupled} was used with a learning rate of 1e-4 and a weight decay of 1e-4. Computations were carried out using NVIDIA GH200 and A100 GPUs for up to 100 epochs.

For the NN method, the temporal window length was set to \(10000 \mu s\). The parameters for the DIF method were set as follows: filter length \(15000 \mu s\), scale 4, and update factor 0.5. The EDnCNN algorithm utilised default parameters and weights provided by the authors in \cite{baldwin2020event}.
\section{Experiments and Results}
\label{sec:experiments}


The objective of this study was to evaluate the impact of varying noise intensity on the performance of deep neural network models and to test the hypothesis that training these networks on input data with different noise levels can improve their effectiveness under changing conditions. This section provides an analysis of the experimental results, comparing the effectiveness of the noise-injection data augmentation method with standard filtering approaches. For additional experiments, including evaluation on a detection task and tests using real camera noise, we refer to the supplementary materials.

\subsection{Methodology}

In order to provide a comprehensive evaluation of the efficacy of the proposed method, four training datasets variants were considered:
(1) \textit{Original} –- the original dataset without any modifications, allowing the assessment of models natural robustness to noise and enabling a comparison with results from other studies that use the same datasets;
(2) \textit{Filtered} –- the training data underwent noise filtering and served as a baseline by providing a dataset with little or no noise, to which generated noise was subsequently added;
(3) \textit{Noise=1Hz/px} –- training data containing constant noise intensity of 1 Hz/px, enabling evaluation of the robustness of models trained at a specific noise level;
(4) \textit{Noise-injection} -– our proposed method, where training data includes various levels of noise, aimed at enhancing classification stability under varying noise intensities.


Initially, we evaluated the performance of models trained on all four variants using data with a range of noise intensities, without applying any filtering, to determine their inherent resilience to noise. Next, we introduce a preprocessing step involving noise filtering, after which all datasets and models were re-evaluated across the same spectrum of noise intensities.

\subsection{Models training}
\label{subsubsec:training}

The classification results for all models and dataset variants, without applying filtering to test sequences with noise injected, are shown in \cref{fig:plots}. 
When testing on noise-free data (0 Hz/px), models that have been trained on the \textit{Filtered} data and the \textit{Noise-injection} method have demonstrated the highest accuracy. For the N-Caltech101 dataset, our \textit{Noise-injection} method demonstrates a 2\% improvement over the \textit{Filtered} variant for the SNN model. For the N-Cars and Mini N-ImageNet datasets, the \textit{Filtered} variant outperforms the \textit{Noise-injection} method by 1\% and 2\%, respectively, also in the SNN models.

Under test conditions that incorporate noise, CNN, ViT, and SNN models trained using the \textit{Noise-injection} method show significantly greater classification stability, even in the presence of substantial noise levels. Conversely, other training methods demonstrate a substantial decline in classification accuracy as noise intensity increases. Utilising training data with constant noise intensity (1 Hz/px) primarily improves results at comparable noise levels during testing but rapidly degrades accuracy at different noise intensities. This effect is particularly evident in the Mini N-ImageNet dataset, where the \textit{Noise=1Hz/px} variant attained the lowest accuracy on the noise-free test set.

An exception to these observations is the model based on the GCN architecture. In this instance, the classification quality gradually decreases with increasing noise, even when employing the \textit{Noise-injection} method. This effect can be attributed to the limited number of neighbours per node in the graph, complicating effective data representation under significant noise conditions. Nevertheless, the \textit{Noise-injection} method consistently attains the highest accuracy among all the training variants analysed for the GCN model, thereby substantiating its overall effectiveness.

The results presented clearly indicate that the proposed \textit{Noise-injection} training method ensures classification stability for CNN, ViT, and SNN models and better results for GCN model across varying input noise levels, surpassing other analysed training approaches. However, it is important to highlight that in the ideal scenario (i.e. absence of noise), this method occasionally demonstrates slightly lower accuracy compared to the \textit{Filtered} method. This reduction -- which does not exceed 1\% on average -- represents the trade-off for increased noise robustness.

\subsection{Evaluation with filtration}

\begin{table*}[htp]
\centering
\resizebox{\textwidth}{!}{%
\begin{tabular}{cc|cccccccc|cccccccc|cccccccc|}
\cline{3-26}
 &
   &
  \multicolumn{8}{c|}{N-Caltech101 \cite{orchard2015converting}} &
  \multicolumn{8}{c|}{N-Cars \cite{sironi2018hats} } &
  \multicolumn{8}{c|}{Mini N-ImageNet \cite{Kim_2021_ICCV}} \\ \cline{3-26} 
 &
   &
  \multicolumn{2}{c}{NN \cite{delbruck2008frame}} &
  \multicolumn{2}{c}{EDnCNN \cite{baldwin2020event}} &
  \multicolumn{2}{c}{DIF \cite{kowalczyk2023interpolation}} &
  \multicolumn{2}{c|}{w/o} &
  \multicolumn{2}{c}{NN \cite{delbruck2008frame}} &
  \multicolumn{2}{c}{EDnCNN \cite{baldwin2020event}} &
  \multicolumn{2}{c}{DIF \cite{kowalczyk2023interpolation}} &
  \multicolumn{2}{c|}{w/o} &
  \multicolumn{2}{c}{NN \cite{delbruck2008frame}} &
  \multicolumn{2}{c}{EDnCNN \cite{baldwin2020event}} &
  \multicolumn{2}{c}{DIF \cite{kowalczyk2023interpolation}} &
  \multicolumn{2}{c|}{w/o} \\ \hline
\multicolumn{1}{|c}{Model} &
  Training Set &
  \begin{tabular}[c]{@{}c@{}}Acc. \\ mean\end{tabular} &
  \begin{tabular}[c]{@{}c@{}}Acc. \\ std.\end{tabular} &
  \begin{tabular}[c]{@{}c@{}}Acc. \\ mean\end{tabular} &
  \begin{tabular}[c]{@{}c@{}}Acc. \\ std.\end{tabular} &
  \begin{tabular}[c]{@{}c@{}}Acc. \\ mean\end{tabular} &
  \begin{tabular}[c]{@{}c@{}}Acc. \\ std.\end{tabular} &
  \begin{tabular}[c]{@{}c@{}}Acc. \\ mean\end{tabular} &
  \begin{tabular}[c]{@{}c@{}}Acc. \\ std.\end{tabular} &
  \begin{tabular}[c]{@{}c@{}}Acc. \\ mean\end{tabular} &
  \begin{tabular}[c]{@{}c@{}}Acc. \\ std.\end{tabular} &
  \begin{tabular}[c]{@{}c@{}}Acc. \\ mean\end{tabular} &
  \begin{tabular}[c]{@{}c@{}}Acc. \\ std.\end{tabular} &
  \begin{tabular}[c]{@{}c@{}}Acc. \\ mean\end{tabular} &
  \begin{tabular}[c]{@{}c@{}}Acc. \\ std.\end{tabular} &
  \begin{tabular}[c]{@{}c@{}}Acc. \\ mean\end{tabular} &
  \begin{tabular}[c]{@{}c@{}}Acc. \\ std.\end{tabular} &
  \begin{tabular}[c]{@{}c@{}}Acc. \\ mean\end{tabular} &
  \begin{tabular}[c]{@{}c@{}}Acc. \\ std.\end{tabular} &
  \begin{tabular}[c]{@{}c@{}}Acc. \\ mean\end{tabular} &
  \begin{tabular}[c]{@{}c@{}}Acc. \\ std.\end{tabular} &
  \begin{tabular}[c]{@{}c@{}}Acc. \\ mean\end{tabular} &
  \begin{tabular}[c]{@{}c@{}}Acc. \\ std.\end{tabular} &
  \begin{tabular}[c]{@{}c@{}}Acc. \\ mean\end{tabular} &
  \begin{tabular}[c]{@{}c@{}}Acc. \\ std.\end{tabular} \\ \hline
\multicolumn{1}{|c}{} &
  Original &
  \cellcolor[HTML]{EFEFEF}59.89 &
  5.36 &
  \cellcolor[HTML]{EFEFEF}59.86 &
  3.95 &
  \cellcolor[HTML]{EFEFEF}63.16 &
  3.47 &
  \cellcolor[HTML]{EFEFEF}52.06 &
  10.77 &
  \cellcolor[HTML]{EFEFEF}68.89 &
  13.64 &
  \cellcolor[HTML]{EFEFEF}79.93 &
  4.78 &
  \cellcolor[HTML]{EFEFEF}74.87 &
  13.76 &
  \cellcolor[HTML]{EFEFEF}58.59 &
  8.68 &
  \cellcolor[HTML]{EFEFEF}37.73 &
  2.03 &
  \cellcolor[HTML]{EFEFEF}15.34 &
  1.28 &
  \cellcolor[HTML]{EFEFEF}35.90 &
  2.83 &
  \cellcolor[HTML]{EFEFEF}35.47 &
  8.34 \\
\multicolumn{1}{|c}{} &
  Filtered &
  \cellcolor[HTML]{EFEFEF}58.35 &
  8.53 &
  \cellcolor[HTML]{EFEFEF}63.77 &
  8.55 &
  \cellcolor[HTML]{EFEFEF}63.67 &
  7.38 &
  \cellcolor[HTML]{EFEFEF}50.13 &
  7.61 &
  \cellcolor[HTML]{EFEFEF}76.07 &
  12.53 &
  \cellcolor[HTML]{EFEFEF}80.91 &
  1.66 &
  \cellcolor[HTML]{EFEFEF}81.76 &
  8.59 &
  \cellcolor[HTML]{EFEFEF}64.82 &
  13.93 &
  \cellcolor[HTML]{EFEFEF}40.96 &
  3.05 &
  \cellcolor[HTML]{EFEFEF}20.96 &
  1.79 &
  \cellcolor[HTML]{EFEFEF}41.36 &
  1.49 &
  \cellcolor[HTML]{EFEFEF}35.91 &
  8.56 \\
\multicolumn{1}{|c}{} &
  Noise=1 Hz &
  \cellcolor[HTML]{EFEFEF}67.15 &
  1.73 &
  \cellcolor[HTML]{EFEFEF}64.02 &
  0.99 &
  \cellcolor[HTML]{EFEFEF}66.38 &
  1.46 &
  \cellcolor[HTML]{EFEFEF}62.13 &
  9.69 &
  \cellcolor[HTML]{EFEFEF}87.45 &
  1.48 &
  \cellcolor[HTML]{EFEFEF}77.94 &
  1.49 &
  \cellcolor[HTML]{EFEFEF}86.70 &
  1.19 &
  \cellcolor[HTML]{EFEFEF}78.90 &
  16.38 &
  \cellcolor[HTML]{EFEFEF}37.56 &
  2.27 &
  \cellcolor[HTML]{EFEFEF}16.46 &
  0.86 &
  \cellcolor[HTML]{EFEFEF}35.73 &
  2.88 &
  \cellcolor[HTML]{EFEFEF}39.00 &
  6.23 \\
\multicolumn{1}{|c}{\multirow{-4}{*}{CNN}} &
  \textbf{Noise-injection} &
  \cellcolor[HTML]{EFEFEF}69.07 &
  0.25 &
  \cellcolor[HTML]{EFEFEF}67.60 &
  0.40 &
  \cellcolor[HTML]{EFEFEF}\underline{69.43} &
  0.21 &
  \cellcolor[HTML]{EFEFEF}\textbf{69.49} &
  0.14 &
  \cellcolor[HTML]{EFEFEF}\underline{89.47} &
  0.32 &
  \cellcolor[HTML]{EFEFEF}82.70 &
  0.83 &
  \cellcolor[HTML]{EFEFEF}88.53 &
  0.61 &
  \cellcolor[HTML]{EFEFEF}\textbf{91.65} &
  0.39 &
  \cellcolor[HTML]{EFEFEF}\underline{43.53} &
  0.51 &
  \cellcolor[HTML]{EFEFEF}22.30 &
  0.88 &
  \cellcolor[HTML]{EFEFEF}43.51 &
  1.17 &
  \cellcolor[HTML]{EFEFEF}\textbf{45.57} &
  0.32 \\ \hline
\multicolumn{1}{|c}{} &
  Original &
  \cellcolor[HTML]{EFEFEF}66.55 &
  0.50 &
  \cellcolor[HTML]{EFEFEF}55.27 &
  5.73 &
  \cellcolor[HTML]{EFEFEF}66.57 &
  0.86 &
  \cellcolor[HTML]{EFEFEF}66.01 &
  2.80 &
  \cellcolor[HTML]{EFEFEF}77.68 &
  8.60 &
  \cellcolor[HTML]{EFEFEF}78.16 &
  1.10 &
  \cellcolor[HTML]{EFEFEF}81.78 &
  7.49 &
  \cellcolor[HTML]{EFEFEF}68.15 &
  10.33 &
  \cellcolor[HTML]{EFEFEF}37.38 &
  2.53 &
  \cellcolor[HTML]{EFEFEF}17.81 &
  0.47 &
  \cellcolor[HTML]{EFEFEF}35.59 &
  1.48 &
  \cellcolor[HTML]{EFEFEF}35.41 &
  6.88 \\
\multicolumn{1}{|c}{} &
  Filtered &
  \cellcolor[HTML]{EFEFEF}69.50 &
  0.82 &
  \cellcolor[HTML]{EFEFEF}59.22 &
  5.76 &
  \cellcolor[HTML]{EFEFEF}69.34 &
  0.61 &
  \cellcolor[HTML]{EFEFEF}67.24 &
  3.40 &
  \cellcolor[HTML]{EFEFEF}81.80 &
  7.19 &
  \cellcolor[HTML]{EFEFEF}82.29 &
  1.54 &
  \cellcolor[HTML]{EFEFEF}84.63 &
  4.85 &
  \cellcolor[HTML]{EFEFEF}73.72 &
  11.54 &
  \cellcolor[HTML]{EFEFEF}41.92 &
  4.53 &
  \cellcolor[HTML]{EFEFEF}23.23 &
  0.94 &
  \cellcolor[HTML]{EFEFEF}42.18 &
  3.13 &
  \cellcolor[HTML]{EFEFEF}37.08 &
  9.00 \\
\multicolumn{1}{|c}{} &
  Noise=1 Hz &
  \cellcolor[HTML]{EFEFEF}67.34 &
  1.01 &
  \cellcolor[HTML]{EFEFEF}56.46 &
  5.81 &
  \cellcolor[HTML]{EFEFEF}66.89 &
  0.86 &
  \cellcolor[HTML]{EFEFEF}68.18 &
  0.87 &
  \cellcolor[HTML]{EFEFEF}86.75 &
  1.76 &
  \cellcolor[HTML]{EFEFEF}80.33 &
  1.00 &
  \cellcolor[HTML]{EFEFEF}87.09 &
  1.07 &
  \cellcolor[HTML]{EFEFEF}79.81 &
  15.26 &
  \cellcolor[HTML]{EFEFEF}37.68 &
  3.41 &
  \cellcolor[HTML]{EFEFEF}16.33 &
  0.70 &
  \cellcolor[HTML]{EFEFEF}35.11 &
  2.83 &
  \cellcolor[HTML]{EFEFEF}40.80 &
  4.94 \\
\multicolumn{1}{|c}{\multirow{-4}{*}{ViT}} &
  \textbf{Noise-injection} &
  \cellcolor[HTML]{EFEFEF}\underline{69.51} &
  0.17 &
  \cellcolor[HTML]{EFEFEF}61.38 &
  5.19 &
  \cellcolor[HTML]{EFEFEF}69.46 &
  0.22 &
  \cellcolor[HTML]{EFEFEF}\textbf{69.57} &
  0.30 &
  \cellcolor[HTML]{EFEFEF}\underline{89.53} &
  0.18 &
  \cellcolor[HTML]{EFEFEF}82.30 &
  1.07 &
  \cellcolor[HTML]{EFEFEF}89.38 &
  0.30 &
  \cellcolor[HTML]{EFEFEF}\textbf{91.19} &
  0.67 &
  \cellcolor[HTML]{EFEFEF}\underline{45.51} &
  0.23 &
  \cellcolor[HTML]{EFEFEF}23.97 &
  0.62 &
  \cellcolor[HTML]{EFEFEF}44.57 &
  0.59 &
  \cellcolor[HTML]{EFEFEF}\textbf{48.04} &
  0.26 \\ \hline
\multicolumn{1}{|c}{} &
  Original &
  \cellcolor[HTML]{EFEFEF}49.57 &
  4.12 &
  \cellcolor[HTML]{EFEFEF}50.80 &
  1.96 &
  \cellcolor[HTML]{EFEFEF}49.85 &
  3.01 &
  \cellcolor[HTML]{EFEFEF}46.45 &
  9.65 &
  \cellcolor[HTML]{EFEFEF}77.04 &
  7.54 &
  \cellcolor[HTML]{EFEFEF}76.20 &
  1.13 &
  \cellcolor[HTML]{EFEFEF}79.53 &
  5.73 &
  \cellcolor[HTML]{EFEFEF}69.36 &
  13.63 &
  \cellcolor[HTML]{EFEFEF}30.37 &
  1.25 &
  \cellcolor[HTML]{EFEFEF}10.73 &
  0.79 &
  \cellcolor[HTML]{EFEFEF}28.77 &
  1.75 &
  \cellcolor[HTML]{EFEFEF}32.22 &
  5.76 \\
\multicolumn{1}{|c}{} &
  Filtered &
  \cellcolor[HTML]{EFEFEF}52.56 &
  10.90 &
  \cellcolor[HTML]{EFEFEF}55.48 &
  6.52 &
  \cellcolor[HTML]{EFEFEF}56.50 &
  6.90 &
  \cellcolor[HTML]{EFEFEF}38.61 &
  16.97 &
  \cellcolor[HTML]{EFEFEF}76.55 &
  12.20 &
  \cellcolor[HTML]{EFEFEF}78.02 &
  0.66 &
  \cellcolor[HTML]{EFEFEF}81.21 &
  7.92 &
  \cellcolor[HTML]{EFEFEF}62.43 &
  15.67 &
  \cellcolor[HTML]{EFEFEF}\underline{38.04} &
  2.20 &
  \cellcolor[HTML]{EFEFEF}15.65 &
  0.88 &
  \cellcolor[HTML]{EFEFEF}37.15 &
  1.69 &
  \cellcolor[HTML]{EFEFEF}34.00 &
  9.60 \\
\multicolumn{1}{|c}{} &
  Noise=1 Hz &
  \cellcolor[HTML]{EFEFEF}59.57 &
  2.81 &
  \cellcolor[HTML]{EFEFEF}54.43 &
  0.79 &
  \cellcolor[HTML]{EFEFEF}57.19 &
  2.55 &
  \cellcolor[HTML]{EFEFEF}55.26 &
  13.25 &
  \cellcolor[HTML]{EFEFEF}84.73 &
  2.41 &
  \cellcolor[HTML]{EFEFEF}72.85 &
  1.34 &
  \cellcolor[HTML]{EFEFEF}83.61 &
  2.36 &
  \cellcolor[HTML]{EFEFEF}81.39 &
  12.89 &
  \cellcolor[HTML]{EFEFEF}30.78 &
  3.00 &
  \cellcolor[HTML]{EFEFEF}10.36 &
  0.54 &
  \cellcolor[HTML]{EFEFEF}28.30 &
  3.27 &
  \cellcolor[HTML]{EFEFEF}33.89 &
  5.73 \\
\multicolumn{1}{|c}{\multirow{-4}{*}{SNN}} &
  \textbf{Noise-injection} &
  \cellcolor[HTML]{EFEFEF}\underline{65.07} &
  0.61 &
  \cellcolor[HTML]{EFEFEF}60.15 &
  2.16 &
  \cellcolor[HTML]{EFEFEF}64.56 &
  0.62 &
  \cellcolor[HTML]{EFEFEF}\textbf{65.77} &
  0.42 &
  \cellcolor[HTML]{EFEFEF}\underline{87.63} &
  0.29 &
  \cellcolor[HTML]{EFEFEF}78.19 &
  1.40 &
  \cellcolor[HTML]{EFEFEF}87.28 &
  0.66 &
  \cellcolor[HTML]{EFEFEF}\textbf{89.37} &
  0.28 &
  \cellcolor[HTML]{EFEFEF}37.00 &
  0.54 &
  \cellcolor[HTML]{EFEFEF}15.48 &
  0.52 &
  \cellcolor[HTML]{EFEFEF}34.97 &
  0.86 &
  \cellcolor[HTML]{EFEFEF}\textbf{40.32} &
  0.85 \\ \hline
\multicolumn{1}{|c}{} &
  Original &
  \cellcolor[HTML]{EFEFEF}51.78 &
  6.57 &
  \cellcolor[HTML]{EFEFEF}46.82 &
  3.90 &
  \cellcolor[HTML]{EFEFEF}54.11 &
  2.48 &
  \cellcolor[HTML]{EFEFEF}41.04 &
  15.21 &
  \cellcolor[HTML]{EFEFEF}73.94 &
  2.51 &
  \cellcolor[HTML]{EFEFEF}66.38 &
  0.62 &
  \cellcolor[HTML]{EFEFEF}75.03 &
  2.26 &
  \cellcolor[HTML]{EFEFEF}65.79 &
  11.85 &
  \cellcolor[HTML]{EFEFEF}37.81 &
  1.26 &
  \cellcolor[HTML]{EFEFEF}15.49 &
  0.32 &
  \cellcolor[HTML]{EFEFEF}37.97 &
  1.16 &
  \cellcolor[HTML]{EFEFEF}36.31 &
  3.06 \\
\multicolumn{1}{|c}{} &
  Filtered &
  \cellcolor[HTML]{EFEFEF}46.74 &
  17.67 &
  \cellcolor[HTML]{EFEFEF}50.69 &
  9.10 &
  \cellcolor[HTML]{EFEFEF}56.66 &
  10.81 &
  \cellcolor[HTML]{EFEFEF}26.92 &
  21.03 &
  \cellcolor[HTML]{EFEFEF}74.76 &
  4.33 &
  \cellcolor[HTML]{EFEFEF}68.15 &
  0.60 &
  \cellcolor[HTML]{EFEFEF}76.04 &
  3.50 &
  \cellcolor[HTML]{EFEFEF}64.76 &
  11.96 &
  \cellcolor[HTML]{EFEFEF}38.57 &
  2.75 &
  \cellcolor[HTML]{EFEFEF}15.97 &
  0.34 &
  \cellcolor[HTML]{EFEFEF}39.18 &
  2.17 &
  \cellcolor[HTML]{EFEFEF}35.31 &
  3.64 \\
\multicolumn{1}{|c}{} &
  Noise=1 Hz &
  \cellcolor[HTML]{EFEFEF}57.07 &
  3.09 &
  \cellcolor[HTML]{EFEFEF}47.55 &
  3.57 &
  \cellcolor[HTML]{EFEFEF}55.21 &
  2.54 &
  \cellcolor[HTML]{EFEFEF}55.76 &
  6.86 &
  \cellcolor[HTML]{EFEFEF}74.26 &
  1.21 &
  \cellcolor[HTML]{EFEFEF}64.85 &
  0.63 &
  \cellcolor[HTML]{EFEFEF}73.71 &
  1.17 &
  \cellcolor[HTML]{EFEFEF}71.03 &
  8.44 &
  \cellcolor[HTML]{EFEFEF}38.71 &
  1.73 &
  \cellcolor[HTML]{EFEFEF}14.40 &
  0.15 &
  \cellcolor[HTML]{EFEFEF}37.69 &
  2.01 &
  \cellcolor[HTML]{EFEFEF}38.79 &
  3.24 \\
\multicolumn{1}{|c}{\multirow{-4}{*}{GCN}} &
  \textbf{Noise-injection} &
  \cellcolor[HTML]{EFEFEF}\underline{59.99} &
  2.18 &
  \cellcolor[HTML]{EFEFEF}52.35 &
  2.54 &
  \cellcolor[HTML]{EFEFEF}\textbf{60.38} &
  1.35 &
  \cellcolor[HTML]{EFEFEF}56.25 &
  6.07 &
  \cellcolor[HTML]{EFEFEF}\underline{76.35} &
  0.86 &
  \cellcolor[HTML]{EFEFEF}66.58 &
  0.39 &
  \cellcolor[HTML]{EFEFEF}\textbf{76.82} &
  0.72 &
  \cellcolor[HTML]{EFEFEF}71.41 &
  8.56 &
  \cellcolor[HTML]{EFEFEF}\underline{41.07} &
  0.73 &
  \cellcolor[HTML]{EFEFEF}16.02 &
  0.41 &
  \cellcolor[HTML]{EFEFEF}40.28 &
  0.86 &
  \cellcolor[HTML]{EFEFEF}\textbf{41.75} &
  0.81 \\ \hline
\end{tabular}%
}
\caption{Quantitative results of different models evaluated on the N-Caltech101 \cite{orchard2015converting}, N-Cars \cite{sironi2018hats}  and Mini N-ImageNet \cite{Kim_2021_ICCV} datasets, using NN \cite{delbruck2008frame}, EDnCNN \cite{baldwin2020event}, DIF \cite{kowalczyk2023interpolation} and without filtration. Results represent the mean and standard deviation of accuracies computed across all noise levels, highlighting the superiority of our proposed method compared to alternative approaches. \textbf{Bold} and \underline{underscore} indicate the highest and second-highest values for each model and dataset, respectively.}

\label{tab:big_table}
\end{table*}

A quantitative evaluation of the impact of state-of-the-art filtering techniques on classification performance for the considered neural networks, along with their comparison to our proposed training method, is presented in \cref{tab:big_table}. 
For the sake of clarity, the table presents averaged results and their standard deviations across the entire noise spectrum. The supplementary material provides detailed results.

The analysis clearly indicates that models trained using our \textit{Noise-injection} method achieve the highest effectiveness across all the considered architectures and datasets. For models that were not exposed to noise during the training process (the \textit{Filtered} variant), the performance was generally enhanced through the filtration of the test data. However, this was not observed in the EDnCNN filter results for the ViT network and the N-Caltech101 dataset. This can be attributed to the fact that the ViT was already able to handle noisy data very effectively in this case, even without disturbances in the training set.

\begin{figure}[!t]
    \centering

    \begin{subfigure}[b]{0.15\textwidth}
        \begin{tikzpicture}[scale=0.33]
        \begin{axis}[
            xlabel={{\Large Noise [Hz/px]}},
            ylabel={{\Large Rate}},
            legend style={font=\Large},
            clip=false,
            every axis y label/.style={
            at={(ticklabel cs:0.5)},rotate=90,anchor=center,
        },
            xmin=0, xmax=5,
            ymin=0, ymax=1,
            legend style={at={(1.7,1.05)}, anchor=south, legend columns=-1},
            xmajorgrids=true,
            ymajorgrids=true,
            grid style=dashed,
        ]
        \addplot[cb-one, mark=*] coordinates {
        (0.01, 0.93)
        (0.05, 0.93)
        (0.1, 0.93)
        (0.25, 0.93)
        (0.5, 0.93)
        (0.75, 0.93)
        (1, 0.93)
        (1.5, 0.94)
        (2, 0.94)
        (2.5, 0.94)
        (3, 0.94)
        (4, 0.94)
        (5, 0.95)};

        \addplot[cb-one, mark=x, mark size=3] coordinates {
        (0.01, 0.15)
        (0.05, 0.16)
        (0.1, 0.16)
        (0.25, 0.17)
        (0.5, 0.19)
        (0.75, 0.20)
        (1, 0.22)
        (1.5, 0.25)
        (2, 0.28)
        (2.5, 0.30)
        (3, 0.33)
        (4, 0.38)
        (5, 0.43)};

        \addplot[cb-two, mark=*] coordinates {
        (0.01, 0.71)
        (0.05, 0.71)
        (0.1, 0.71)
        (0.25, 0.71)
        (0.5, 0.72)
        (0.75, 0.72)
        (1, 0.72)
        (1.5, 0.73)
        (2, 0.74)
        (2.5, 0.74)
        (3, 0.75)
        (4, 0.76)
        (5, 0.77)};

        \addplot[cb-two, mark=x, mark size=3] coordinates {
        (0.01, 0.14)
        (0.05, 0.14)
        (0.1, 0.15)
        (0.25, 0.15)
        (0.5, 0.15)
        (0.75, 0.15)
        (1, 0.16)
        (1.5, 0.17)
        (2, 0.18)
        (2.5, 0.20)
        (3, 0.22)
        (4, 0.28)
        (5, 0.33)};

        \addplot[cb-four, mark=*] coordinates {
        (0.01, 0.93)
        (0.05, 0.93)
        (0.1, 0.93)
        (0.25, 0.93)
        (0.5, 0.93)
        (0.75, 0.93)
        (1, 0.94)
        (1.5, 0.94)
        (2, 0.94)
        (2.5, 0.95)
        (3, 0.95)
        (4, 0.96)
        (5, 0.97)};

        \addplot[cb-four, mark=x, mark size=3] coordinates {
        (0.01, 0.21)
        (0.05, 0.21)
        (0.1, 0.21)
        (0.25, 0.21)
        (0.5, 0.22)
        (0.75, 0.22)
        (1, 0.22)
        (1.5, 0.23)
        (2, 0.24)
        (2.5, 0.26)
        (3, 0.28)
        (4, 0.35)
        (5, 0.44)};
        
        \legend{NN\_TPR~~~~~~~~, NN\_FPR~~~~~~~~, EDnCNN\_TPR~~~~~~~~, EDnCNN\_FPR~~~~~~~~, DIF\_TPR~~~~~~~~, DIF\_FPR~~~~~~~~}
        
        \end{axis}
        \end{tikzpicture}
        \caption{N-Caltech101}
    \end{subfigure}
    \begin{subfigure}[b]{0.15\textwidth }
        \begin{tikzpicture}[scale=0.33]
        \begin{axis}[
            xlabel={{\Large Noise [Hz/px]}},
            ylabel={{\Large Rate}},
            clip=false,
            every axis y label/.style={
            at={(ticklabel cs:0.5)},rotate=90,anchor=center,
        },
            xmin=0, xmax=5,
            ymin=0, ymax=1,
            xmajorgrids=true,
            ymajorgrids=true,
            grid style=dashed,
        ]
        \addplot[cb-one, mark=*] coordinates {
        (0.01, 0.73)
        (0.05, 0.73)
        (0.1, 0.73)
        (0.25, 0.74)
        (0.5, 0.74)
        (0.75, 0.75)
        (1, 0.75)
        (1.5, 0.76)
        (2, 0.77)
        (2.5, 0.78)
        (3, 0.79)
        (4, 0.80)
        (5, 0.81)};

        \addplot[cb-one, mark=x, mark size=3] coordinates {
        (0.01, 0.09)
        (0.05, 0.09)
        (0.1, 0.09)
        (0.25, 0.10)
        (0.5, 0.12)
        (0.75, 0.14)
        (1, 0.15)
        (1.5, 0.18)
        (2, 0.21)
        (2.5, 0.24)
        (3, 0.27)
        (4, 0.32)
        (5, 0.37)};

        \addplot[cb-two, mark=*] coordinates {
        (0.01, 0.22)
        (0.05, 0.22)
        (0.1, 0.22)
        (0.25, 0.22)
        (0.5, 0.23)
        (0.75, 0.23)
        (1, 0.23)
        (1.5, 0.23)
        (2, 0.24)
        (2.5, 0.24)
        (3, 0.25)
        (4, 0.26)
        (5, 0.27)};

        \addplot[cb-two, mark=x, mark size=3] coordinates {
        (0.01, 0.02)
        (0.05, 0.02)
        (0.1, 0.02)
        (0.25, 0.02)
        (0.5, 0.02)
        (0.75, 0.02)
        (1, 0.02)
        (1.5, 0.02)
        (2, 0.02)
        (2.5, 0.02)
        (3, 0.02)
        (4, 0.03)
        (5, 0.03)};

        \addplot[cb-four, mark=*] coordinates {
        (0.01, 0.76)
        (0.05, 0.76)
        (0.1, 0.76)
        (0.25, 0.77)
        (0.5, 0.77)
        (0.75, 0.78)
        (1, 0.79)
        (1.5, 0.80)
        (2, 0.82)
        (2.5, 0.83)
        (3, 0.85)
        (4, 0.88)
        (5, 0.91)};

        \addplot[cb-four, mark=x, mark size=3] coordinates {
        (0.01, 0.11)
        (0.05, 0.12)
        (0.1, 0.12)
        (0.25, 0.12)
        (0.5, 0.12)
        (0.75, 0.12)
        (1, 0.13)
        (1.5, 0.14)
        (2, 0.15)
        (2.5, 0.16)
        (3, 0.19)
        (4, 0.25)
        (5, 0.34)};

        \end{axis}
        \end{tikzpicture}
        \caption{N-Cars}
    \end{subfigure}
    \begin{subfigure}[b]{0.15\textwidth}
        \begin{tikzpicture}[scale=0.33]
        \begin{axis}[
            xlabel={{\Large Noise [Hz/px]}},
            ylabel={{\Large Rate}},
            legend style={font=\Large},
            clip=false,
            every axis y label/.style={
            at={(ticklabel cs:0.5)},rotate=90,anchor=center,
        },
            xmin=0, xmax=5,
            ymin=0, ymax=1,
            xmajorgrids=true,
            ymajorgrids=true,
            grid style=dashed,
        ]
        \addplot[cb-one, mark=*] coordinates {
        (0.01, 0.78)
        (0.05, 0.78)
        (0.1, 0.78)
        (0.25, 0.78)
        (0.5, 0.79)
        (0.75, 0.79)
        (1, 0.80)
        (1.5, 0.80)
        (2, 0.81)
        (2.5, 0.82)
        (3, 0.82)
        (4, 0.83)
        (5, 0.85)
        };

        \addplot[cb-one, mark=x, mark size=3] coordinates {
        (0.01, 0.17)
        (0.05, 0.17)
        (0.1, 0.18)
        (0.25, 0.19)
        (0.5, 0.20)
        (0.75, 0.21)
        (1, 0.23)
        (1.5, 0.25)
        (2, 0.28)
        (2.5, 0.31)
        (3, 0.33)
        (4, 0.37)
        (5, 0.41)
        };
        
        \addplot[cb-two, mark=*] coordinates {
        (0.01, 0.21)
        (0.05, 0.21)
        (0.1, 0.21)
        (0.25, 0.21)
        (0.5, 0.21)
        (0.75, 0.21)
        (1, 0.21)
        (1.5, 0.21)
        (2, 0.21)
        (2.5, 0.22)
        (3, 0.22)
        (4, 0.22)
        (5, 0.23)
        };

        \addplot[cb-two, mark=x, mark size=3] coordinates {
        (0.01, 0.03)
        (0.05, 0.03)
        (0.1, 0.03)
        (0.25, 0.03)
        (0.5, 0.03)
        (0.75, 0.03)
        (1, 0.03)
        (1.5, 0.03)
        (2, 0.03)
        (2.5, 0.03)
        (3, 0.03)
        (4, 0.03)
        (5, 0.03)
        };

        \addplot[cb-four, mark=*] coordinates {
        (0.01, 0.80)
        (0.05, 0.80)
        (0.1, 0.80)
        (0.25, 0.80)
        (0.5, 0.81)
        (0.75, 0.82)
        (1, 0.82)
        (1.5, 0.84)
        (2, 0.85)
        (2.5, 0.86)
        (3, 0.87)
        (4, 0.90)
        (5, 0.92)};

        \addplot[cb-four, mark=x, mark size=3] coordinates {
        (0.01, 0.21)
        (0.05, 0.21)
        (0.1, 0.21)
        (0.25, 0.21)
        (0.5, 0.22)
        (0.75, 0.23)
        (1, 0.23)
        (1.5, 0.25)
        (2, 0.27)
        (2.5, 0.29)
        (3, 0.32)
        (4, 0.39)
        (5, 0.46)};

        \end{axis}
        \end{tikzpicture}
        \caption{N-ImageNet}
    \end{subfigure}

    \caption{True-positive and false-positive rates for the NN, EDnCNN and DIF filtering methods for the test sequences of the data sets used.}
\label{fig:plots_tpr_fpr}
\end{figure}
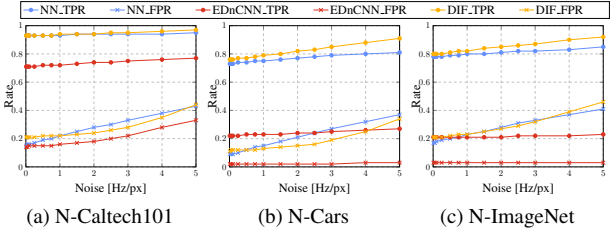

Furthermore, it is noteworthy that for CNN, ViT, and SNN models, the application of any event filtering alongside our proposed method led to a decline in classification quality. The particularly negative impact observed with the \textit{EDnCNN} method can be attributed to its aggressive filtering nature, leading to excessive reduction of input events. This observation is confirmed by \cref{fig:plots_tpr_fpr}, which shows true-positive and false-positive rates for different noise intensities in the test data across the evaluated filtering methods. It can be observed that the \textit{EDnCNN} removes substantially more events than the other tested methods.
The TPR and FPR plots presented also demonstrate that the performance of filtering algorithms varies depending on the noise intensity. Consequently, it is not feasible to expect the same level of accuracy for the neural network under different environmental conditions. In order to get the best results, the filtering parameters should be selected according to the operating conditions of the event sensor, which would require changing its parameters during system operation, thus implementing adaptability.

An exception, once again, is the GCN architecture, which, although achieving the highest performance with the \textit{Noise-injection} method, also benefits from additional filtering using the \textit{DIF} or \textit{NN} methods. This finding suggests that graph-based models exhibit particular sensitivity to the number of input events, and that moderate filtering can enhance data representation effectiveness. At the same time, these results demonstrate that, even for the GCN architecture, our proposed \textit{Noise-injection} method significantly enhances classification quality, particularly in conjunction with filtering techniques.

\subsection{Qualitative results}

\begin{figure*}[htp]
    \centering

    \begin{subfigure}{0.01\textwidth}
        \centering
        \rotatebox{90}{~{\tiny Events}}
    \end{subfigure}
    \begin{subfigure}{0.07\textwidth}
        \centering
        \includegraphics[width=\textwidth]{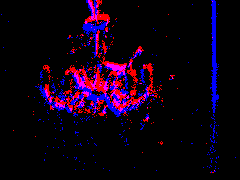}
    \end{subfigure}
    \begin{subfigure}{0.07\textwidth}
        \centering
        \includegraphics[width=\textwidth]{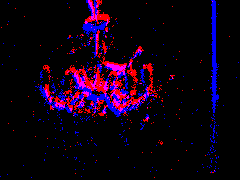}
    \end{subfigure}
    \begin{subfigure}{0.07\textwidth}
        \centering
        \includegraphics[width=\textwidth]{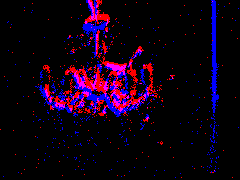}
    \end{subfigure}
    \begin{subfigure}{0.07\textwidth}
        \centering
        \includegraphics[width=\textwidth]{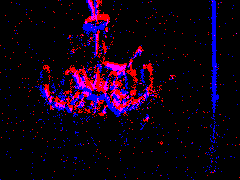}
    \end{subfigure}
    \begin{subfigure}{0.07\textwidth}
        \centering
        \includegraphics[width=\textwidth]{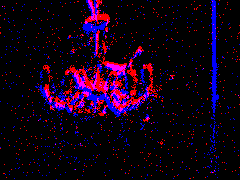}
    \end{subfigure}
    \begin{subfigure}{0.07\textwidth}
        \centering
        \includegraphics[width=\textwidth]{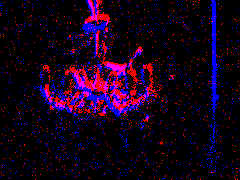}
    \end{subfigure}
    \begin{subfigure}{0.07\textwidth}
        \centering
        \includegraphics[width=\textwidth]{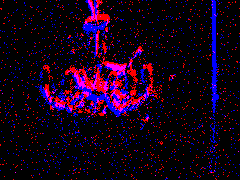}
    \end{subfigure}
    \begin{subfigure}{0.07\textwidth}
        \centering
        \includegraphics[width=\textwidth]{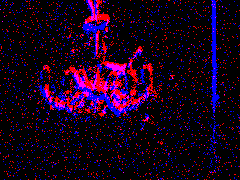}
    \end{subfigure}
    \begin{subfigure}{0.07\textwidth}
        \centering
        \includegraphics[width=\textwidth]{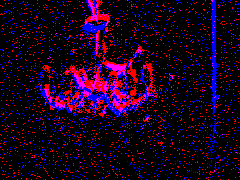}
    \end{subfigure}
    \begin{subfigure}{0.07\textwidth}
        \centering
        \includegraphics[width=\textwidth]{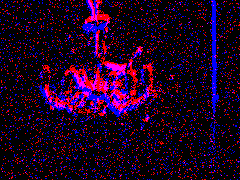}
    \end{subfigure}
    \begin{subfigure}{0.07\textwidth}
        \centering
        \includegraphics[width=\textwidth]{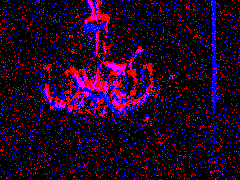}
    \end{subfigure}
    \begin{subfigure}{0.07\textwidth}
        \centering
        \includegraphics[width=\textwidth]{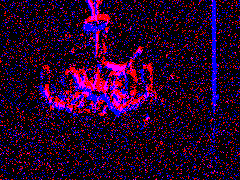}
    \end{subfigure}
    \begin{subfigure}{0.07\textwidth}
        \centering
        \includegraphics[width=\textwidth]{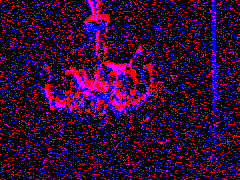}
    \end{subfigure}

    \vspace{2pt}

    \begin{subfigure}{0.01\textwidth}
        \centering
        \rotatebox{90}{~~{\tiny w/o}}
    \end{subfigure}
    \begin{subfigure}{0.07\textwidth}
        \centering
        \includegraphics[width=\textwidth]{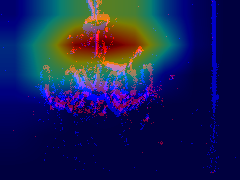}
    \end{subfigure}
    \begin{subfigure}{0.07\textwidth}
        \centering
        \includegraphics[width=\textwidth]{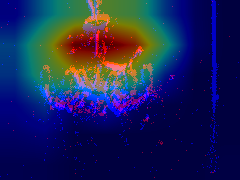}
    \end{subfigure}
    \begin{subfigure}{0.07\textwidth}
        \centering
        \includegraphics[width=\textwidth]{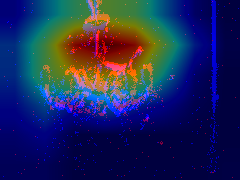}
    \end{subfigure}
    \begin{subfigure}{0.07\textwidth}
        \centering
        \includegraphics[width=\textwidth]{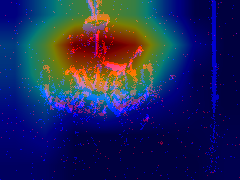}
    \end{subfigure}
    \begin{subfigure}{0.07\textwidth}
        \centering
        \includegraphics[width=\textwidth]{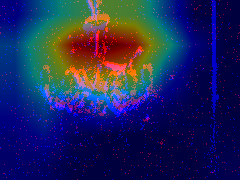}
    \end{subfigure}
    \begin{subfigure}{0.07\textwidth}
        \centering
        \includegraphics[width=\textwidth]{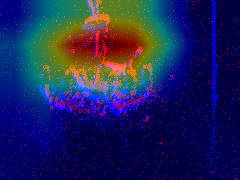}
    \end{subfigure}
    \begin{subfigure}{0.07\textwidth}
        \centering
        \includegraphics[width=\textwidth]{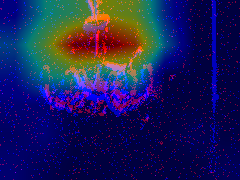}
    \end{subfigure}
    \begin{subfigure}{0.07\textwidth}
        \centering
        \includegraphics[width=\textwidth]{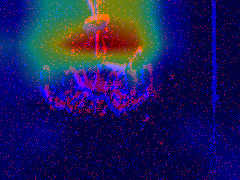}
    \end{subfigure}
    \begin{subfigure}{0.07\textwidth}
        \centering
        \includegraphics[width=\textwidth]{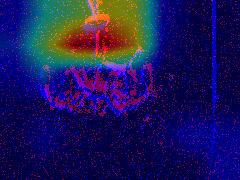}
    \end{subfigure}
    \begin{subfigure}{0.07\textwidth}
        \centering
        \includegraphics[width=\textwidth]{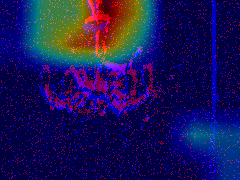}
    \end{subfigure}
    \begin{subfigure}{0.07\textwidth}
        \centering
        \includegraphics[width=\textwidth]{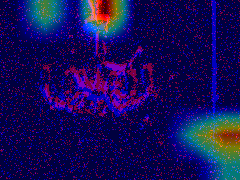}
    \end{subfigure}
    \begin{subfigure}{0.07\textwidth}
        \centering
        \includegraphics[width=\textwidth]{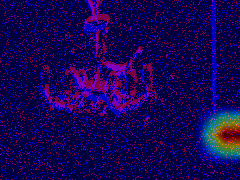}
    \end{subfigure}
    \begin{subfigure}{0.07\textwidth}
        \centering
        \includegraphics[width=\textwidth]{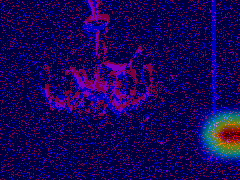}
    \end{subfigure}

    \vspace{2pt}

    \begin{subfigure}{0.01\textwidth}
        \centering
        \rotatebox{90}{~~~{\tiny NN}}
    \end{subfigure}
    \begin{subfigure}{0.07\textwidth}
        \centering
        \includegraphics[width=\textwidth]{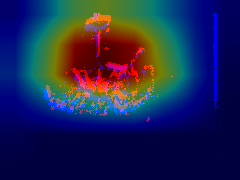}
    \end{subfigure}
    \begin{subfigure}{0.07\textwidth}
        \centering
        \includegraphics[width=\textwidth]{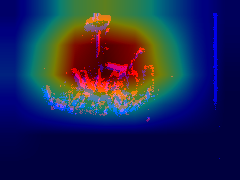}
    \end{subfigure}
    \begin{subfigure}{0.07\textwidth}
        \centering
        \includegraphics[width=\textwidth]{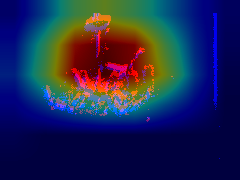}
    \end{subfigure}
    \begin{subfigure}{0.07\textwidth}
        \centering
        \includegraphics[width=\textwidth]{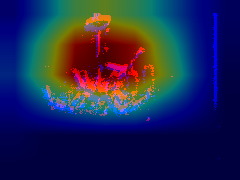}
    \end{subfigure}
    \begin{subfigure}{0.07\textwidth}
        \centering
        \includegraphics[width=\textwidth]{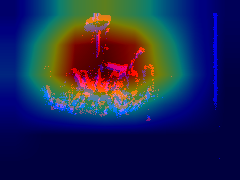}
    \end{subfigure}
    \begin{subfigure}{0.07\textwidth}
        \centering
        \includegraphics[width=\textwidth]{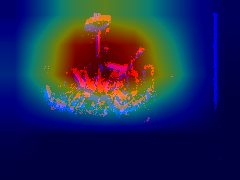}
    \end{subfigure}
    \begin{subfigure}{0.07\textwidth}
        \centering
        \includegraphics[width=\textwidth]{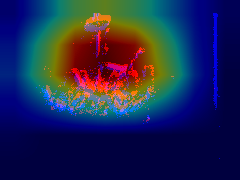}
    \end{subfigure}
    \begin{subfigure}{0.07\textwidth}
        \centering
        \includegraphics[width=\textwidth]{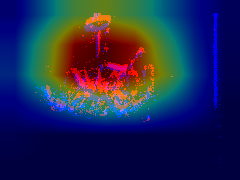}
    \end{subfigure}
    \begin{subfigure}{0.07\textwidth}
        \centering
        \includegraphics[width=\textwidth]{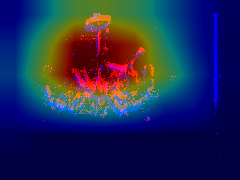}
    \end{subfigure}
    \begin{subfigure}{0.07\textwidth}
        \centering
        \includegraphics[width=\textwidth]{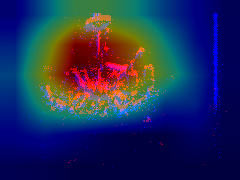}
    \end{subfigure}
    \begin{subfigure}{0.07\textwidth}
        \centering
        \includegraphics[width=\textwidth]{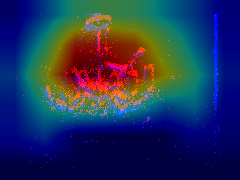}
    \end{subfigure}
    \begin{subfigure}{0.07\textwidth}
        \centering
        \includegraphics[width=\textwidth]{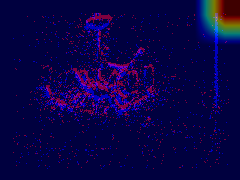}
    \end{subfigure}
    \begin{subfigure}{0.07\textwidth}
        \centering
        \includegraphics[width=\textwidth]{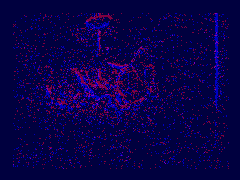}
    \end{subfigure}

    \vspace{2pt}
    
    \begin{subfigure}{0.01\textwidth}
        \centering
        \rotatebox{90}{~~~~~~~{\tiny Ours}}
    \end{subfigure}
    \begin{subfigure}{0.07\textwidth}
        \centering
        \includegraphics[width=\textwidth]{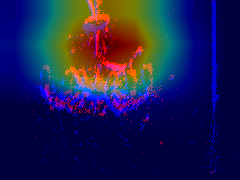}
        \caption*{0.01 Hz}
    \end{subfigure}
    \begin{subfigure}{0.07\textwidth}
        \centering
        \includegraphics[width=\textwidth]{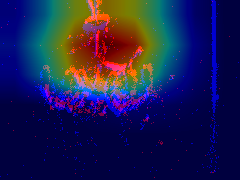}
        \caption*{0.05 Hz}
    \end{subfigure}
    \begin{subfigure}{0.07\textwidth}
        \centering
        \includegraphics[width=\textwidth]{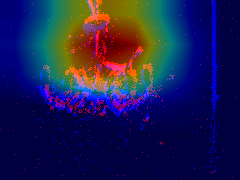}
        \caption*{0.1 Hz}
    \end{subfigure}
    \begin{subfigure}{0.07\textwidth}
        \centering
        \includegraphics[width=\textwidth]{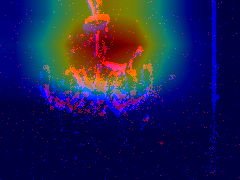}
        \caption*{0.25 Hz}
    \end{subfigure}
    \begin{subfigure}{0.07\textwidth}
        \centering
        \includegraphics[width=\textwidth]{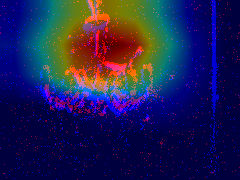}
        \caption*{0.5 Hz}
    \end{subfigure}
    \begin{subfigure}{0.07\textwidth}
        \centering
        \includegraphics[width=\textwidth]{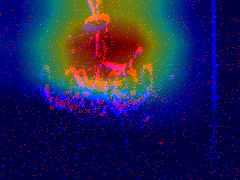}
        \caption*{0.75 Hz}
    \end{subfigure}
    \begin{subfigure}{0.07\textwidth}
        \centering
        \includegraphics[width=\textwidth]{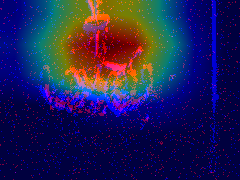}
        \caption*{1 Hz}
    \end{subfigure}
    \begin{subfigure}{0.07\textwidth}
        \centering
        \includegraphics[width=\textwidth]{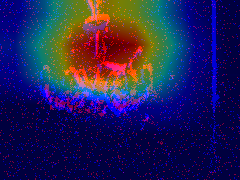}
        \caption*{1.5 Hz}
    \end{subfigure}
    \begin{subfigure}{0.07\textwidth}
        \centering
        \includegraphics[width=\textwidth]{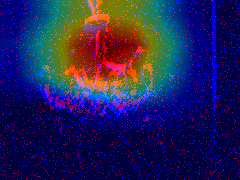}
        \caption*{2 Hz}
    \end{subfigure}
    \begin{subfigure}{0.07\textwidth}
        \centering
        \includegraphics[width=\textwidth]{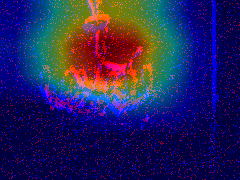}
        \caption*{2.5 Hz}
    \end{subfigure}
    \begin{subfigure}{0.07\textwidth}
        \centering
        \includegraphics[width=\textwidth]{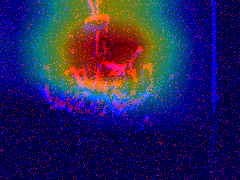}
        \caption*{3 Hz}
    \end{subfigure}
    \begin{subfigure}{0.07\textwidth}
        \centering
        \includegraphics[width=\textwidth]{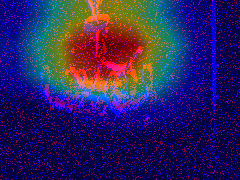}
        \caption*{4 Hz}
    \end{subfigure}
    \begin{subfigure}{0.07\textwidth}
        \centering
        \includegraphics[width=\textwidth]{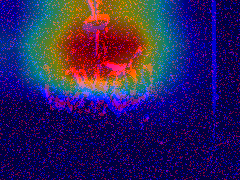}
        \caption*{5 Hz}
    \end{subfigure}

\caption{Qualitative comparison of models trained with our noise-injection method, with an NN-based filter, and without any method across different noise levels. The presented example illustrate the most common effects of noise on activations: the fading of object activations leading to complete loss of activation.}
\label{fig:figures}
\end{figure*}

\cref{fig:figures} presents a visual analysis of N-Caltech101 results obtained using Class Activation Mapping (Grad-CAM) \cite{selvaraju2020grad} for the CNN network trained using our \textit{Noise-injection} method, without it, and after applying \textit{NN} filtration. For more examples on N-Caltech101 and Mini N-ImageNet and with GradCam and GradCam++, we suggest checking the supplementary material. For models trained without the \textit{Noise-injection} augmentation, we observe a decrease in quality as noise levels increase -- the class activations start covering irrelevant areas of the image, and the network frequently ignores key visual features of the object or focuses on noisy background regions. Data filtration can reduce this effect but cannot completely eliminate it. Conversely, the network trained using our method effectively identifies important object features even under increasing noise conditions, maintaining activation stability and accurate target classification. These results confirm the advantage of the proposed \textit{Noise-injection} approach in enhancing neural network robustness against varying levels of noise.

\section{Discussion and Limitations}
\label{sec:discusion}

The quantitative and qualitative analyses conducted in this study confirm the effectiveness of the proposed \textit{Noise-injection} approach for the classification of noisy event-based data. The obtained results suggest that it could also serve as an alternative to filtering data prior to feeding it into neural networks. The models trained using our method achieved the highest average classification accuracy across all analysed architectures and datasets.

A significant advantage of the \textit{Noise-injection} method is its ability to preserve crucial event information that is often lost with more aggressive filtering techniques. Intensive event elimination frequently results in the loss of key visual features, causing a substantial drop in classification quality. In contrast, the \textit{Noise-injection} approach enhances network robustness through exposure to varying levels of noise.

However, these results do not negate the necessity for continued development and research into event data filtering methods. Besides enhancing the efficacy of subsequent algorithms, data filtration reduces the number of events required for processing. This, in turn, can lead to reduced data transmission rates, decreased storage requirements, lower power consumption and computational demand.

Moreover, specific limitations were observed for graph-based neural network models. For these models, even when employing the proposed training method, classification quality gradually decreases with increasing noise levels. This suggests that the effectiveness of our approach for graph models could be further improved by combining it with moderate filtering techniques, such as the \textit{Nearest Neighbours} method, which reduces noise events without excessively losing essential information.

Furthermore, although the \textit{Noise-injection} method significantly enhances model robustness to noise, a slight reduction in classification accuracy was observed under ideal conditions (i.e. completely noise-free scenarios) compared to models trained on \textit{Filtered} data. This reduction can be interpreted as a natural trade-off between maximal accuracy in ideal scenarios and enhanced stability and robustness under realistic, noisy conditions.

\section{Conclusion}
\label{sec:conclusion}




In this paper, we present a novel data augmentation method that involves the controlled injection of noise into event streams. This approach aims to enhance the robustness of models against noise and improve their generalisation capabilities. We validated this approach on classification tasks using different datasets and models, comparing our results with three filtering methods. The experiments demonstrated that our method outperforms traditional filtering techniques by improving the stability of CNN, ViT, and SNN models and achieving superior performance for GCN models. This supports our hypothesis that the introduction of noise has a positive impact on model generalisation.

We believe that the presented method can significantly increase the performance of vision systems by improving the efficiency of neural networks under diverse environmental conditions. The proposed solution has the potential to serve as an alternative to the event data filtering process, which can introduce additional delays and remove some of the relevant events.
Although our experiments were limited to classification tasks, we believe that the proposed approach could also positively impact other tasks realised with the use of neural networks, like object detection.

Future research directions include exploring other types of noise and integrating our method with existing filtering techniques to achieve further improvements. Additionally, we plan to employ knowledge transfer techniques between models trained with and without noise augmentation and focus on solving problems with graph-based networks.

\section{Acknowledgments}

The work presented in this paper was supported by the National Science Centre project no. 2021/41/N/ST6/03915 (first author) and no. 2024/53/N/ST6/04254 (second author). We also gratefully acknowledge Polish high-performance computing infrastructure PLGrid (HPC Center: ACK Cyfronet AGH) for providing computer facilities and support within computational grant no. PLG/2024/017438 (first author) and PLG/2025/017956 (second author).
{
    \small
    \bibliographystyle{ieeenat_fullname}
    \bibliography{main}
}

\clearpage
\setcounter{page}{1}
\maketitlesupplementary

\section{Detection results}

In order to examine the impact of our method on another, more complex task, we used the N-Caltech101 dataset for object detection. For this purpose, we employed YOLOX as the detection head and exactly the same CNN and ViT architectures as in the classification study (ResNet18 and MaxViT – detailed information can be found in Table \ref{table:models} in the main part of the work).

\begin{table*}[!t]
\resizebox{\textwidth}{!}{%
\begin{tabular}{@{}cc|llllllllllll|@{}}
\cmidrule(l){3-14}
 &
   &
  \multicolumn{3}{c}{NN} &
  \multicolumn{3}{c}{EDnCNN} &
  \multicolumn{3}{c}{DIF} &
  \multicolumn{3}{c|}{w/o} \\ \midrule
\multicolumn{1}{|c}{Model} &
  Training Set &
  \multicolumn{1}{c}{mAP} &
  \multicolumn{1}{c}{mAP@50} &
  \multicolumn{1}{c}{mAP@75} &
  \multicolumn{1}{c}{mAP} &
  \multicolumn{1}{c}{mAP@50} &
  \multicolumn{1}{c}{mAP@75} &
  \multicolumn{1}{c}{mAP} &
  \multicolumn{1}{c}{mAP@50} &
  \multicolumn{1}{c}{mAP@75} &
  \multicolumn{1}{c}{mAP} &
  \multicolumn{1}{c}{mAP@50} &
  \multicolumn{1}{c|}{mAP@75} \\ \midrule
\multicolumn{1}{|c}{} &
  Original &
  \cellcolor[HTML]{EFEFEF}30.28 \(\pm\) 2.22 &
  57.57 \(\pm\) 1.43 &
  \cellcolor[HTML]{EFEFEF}27.80 \(\pm\) 3.89 &
  31.23 \(\pm\) 1.51 &
  \cellcolor[HTML]{EFEFEF}56.42 \(\pm\) 1.57 &
  30.17 \(\pm\) 2.44 &
  \cellcolor[HTML]{EFEFEF}31.50 \(\pm\) 1.92 &
  58.04 \(\pm\) 1.22 &
  \cellcolor[HTML]{EFEFEF}30.16 \(\pm\) 3.53 &
  28.62 \(\pm\) 1.80 &
  \cellcolor[HTML]{EFEFEF}56.56 \(\pm\) 2.18 &
  24.91 \(\pm\) 2.66 \\
\multicolumn{1}{|c}{} &
  Filtered &
  \cellcolor[HTML]{EFEFEF}29.29 \(\pm\) 1.46 &
  55.62 \(\pm\) 1.66 &
  \cellcolor[HTML]{EFEFEF}27.36 \(\pm\) 1.79 &
  29.81 \(\pm\) 1.99 &
  \cellcolor[HTML]{EFEFEF}54.91 \(\pm\) 2.88 &
  28.56 \(\pm\) 2.75 &
  \cellcolor[HTML]{EFEFEF}30.00 \(\pm\) 1.51 &
  56.18 \(\pm\) 1.30 &
  \cellcolor[HTML]{EFEFEF}27.96 \(\pm\) 2.17 &
  28.12 \(\pm\) 1.33 &
  \cellcolor[HTML]{EFEFEF}53.66 \(\pm\) 2.42 &
  26.07 \(\pm\) 1.22 \\
\multicolumn{1}{|c}{} &
  Noise=1Hz &
  \cellcolor[HTML]{EFEFEF}27.81 \(\pm\) 5.06 &
  53.74 \(\pm\) 6.26 &
  \cellcolor[HTML]{EFEFEF}25.13 \(\pm\) 6.13 &
  24.13 \(\pm\) 3.29 &
  \cellcolor[HTML]{EFEFEF}48.16 \(\pm\) 4.70 &
  20.71 \(\pm\) 4.31 &
  \cellcolor[HTML]{EFEFEF}24.44 \(\pm\) 3.84 &
  49.60 \(\pm\) 5.07 &
  \cellcolor[HTML]{EFEFEF}21.13 \(\pm\) 4.52 &
  30.54 \(\pm\) 3.85 &
  \cellcolor[HTML]{EFEFEF}57.50 \(\pm\) 4.74 &
  28.18 \(\pm\) 4.88 \\
\multicolumn{1}{|c}{\multirow{-4}{*}{CNN}} &
  \textbf{Noise-injection} &
  \cellcolor[HTML]{EFEFEF}34.06 \(\pm\) 1.10 &
  60.40 \(\pm\) 1.10 &
  \cellcolor[HTML]{EFEFEF}34.11 \(\pm\) 1.95 &
  32.14 \(\pm\) 0.59 &
  \cellcolor[HTML]{EFEFEF}58.20 \(\pm\) 0.67 &
  30.78 \(\pm\) 1.07 &
  \cellcolor[HTML]{EFEFEF}33.08 \(\pm\) 1.10 &
  59.48 \(\pm\) 1.19 &
  \cellcolor[HTML]{EFEFEF}32.38 \(\pm\) 1.79 &
  \textbf{34.94 \(\pm\) 0.43} &
  \cellcolor[HTML]{EFEFEF}\textbf{61.17 \(\pm\) 0.33} &
  \textbf{35.60 \(\pm\) 0.78} \\ \midrule
\multicolumn{1}{|c}{} &
  Original &
  \cellcolor[HTML]{EFEFEF}20.56 \(\pm\) 4.25 &
  40.34 \(\pm\) 5.01 &
  \cellcolor[HTML]{EFEFEF}18.22 \(\pm\) 5.56 &
  17.46 \(\pm\) 4.08 &
  \cellcolor[HTML]{EFEFEF}32.80 \(\pm\) 6.93 &
  15.55 \(\pm\) 3.96 &
  \cellcolor[HTML]{EFEFEF}22.86 \(\pm\) 2.84 &
  43.03 \(\pm\) 3.41 &
  \cellcolor[HTML]{EFEFEF}21.20 \(\pm\) 3.85 &
  16.82 \(\pm\) 4.66 &
  \cellcolor[HTML]{EFEFEF}35.82 \(\pm\) 6.74 &
  13.30 \(\pm\) 5.65 \\
\multicolumn{1}{|c}{} &
  Filtered &
  \cellcolor[HTML]{EFEFEF}25.55 \(\pm\) 4.21 &
  48.53 \(\pm\) 4.56 &
  \cellcolor[HTML]{EFEFEF}23.57 \(\pm\) 5.98 &
  22.86 \(\pm\) 5.48 &
  \cellcolor[HTML]{EFEFEF}41.27 \(\pm\) 9.19 &
  21.94 \(\pm\) 5.74 &
  \cellcolor[HTML]{EFEFEF}27.63 \(\pm\) 3.18 &
  50.31 \(\pm\) 3.45 &
  \cellcolor[HTML]{EFEFEF}26.71 \(\pm\) 4.46 &
  19.95 \(\pm\) 5.51 &
  \cellcolor[HTML]{EFEFEF}41.52 \(\pm\) 7.96 &
  16.07 \(\pm\) 6.98 \\
\multicolumn{1}{|c}{} &
  Noise=1Hz &
  \cellcolor[HTML]{EFEFEF}27.15 \(\pm\) 4.37 &
  50.50 \(\pm\) 4.62 &
  \cellcolor[HTML]{EFEFEF}25.00 \(\pm\) 6.25 &
  17.65 \(\pm\) 0.83 &
  \cellcolor[HTML]{EFEFEF}34.66 \(\pm\) 2.58 &
  15.72 \(\pm\) 0.77 &
  \cellcolor[HTML]{EFEFEF}24.57 \(\pm\) 3.28 &
  47.44 \(\pm\) 3.96 &
  \cellcolor[HTML]{EFEFEF}21.21 \(\pm\) 4.69 &
  29.67 \(\pm\) 3.85 &
  \cellcolor[HTML]{EFEFEF}54.43 \(\pm\) 3.78 &
  28.53 \(\pm\) 5.89 \\
\multicolumn{1}{|c}{\multirow{-4}{*}{ViT}} &
  \textbf{Noise-injection} &
  \cellcolor[HTML]{EFEFEF}31.12 \(\pm\) 1.26 &
  54.66 \(\pm\) 1.08 &
  \cellcolor[HTML]{EFEFEF}31.01 \(\pm\) 1.63 &
  23.47 \(\pm\) 2.22 &
  \cellcolor[HTML]{EFEFEF}41.86 \(\pm\) 4.23 &
  22.74 \(\pm\) 2.18 &
  \cellcolor[HTML]{EFEFEF}30.44 \(\pm\) 1.11 &
  54.16 \(\pm\) 1.25 &
  \cellcolor[HTML]{EFEFEF}30.04 \(\pm\) 1.58 &
  \textbf{32.06 \(\pm\) 0.61} &
  \cellcolor[HTML]{EFEFEF}\textbf{55.65 \(\pm\) 0.56} &
  \textbf{32.14 \(\pm\) 0.92} \\ \bottomrule
\end{tabular}%
}
\caption{Quantitative results for CNN and ViT model on the N-Caltech101 datasets in detection problem using NN, EDnCNN, DIF and without filtration. Results represents the mean and standard deviation of mAP (mean average precision). \textbf{Bold} indicate the highest values for each model.}
\label{table:detection}
\end{table*}
\input{sec/detection_training_plots}
\begin{figure}[!t]
    \centering

    \begin{subfigure}[b]{0.2\textwidth}
        \begin{tikzpicture}[scale=0.45]
	    \begin{axis}[
		xlabel={Threshold},
        clip=false,
		ylabel={Rate},
		legend pos=south east,
        xmajorgrids=true,
        ymajorgrids=true,
        grid style=dashed,
	    ]

	    \addplot [
		blue,
        dashed,
		forget plot
	    ] table [
		x=Threshold,
		y expr=\thisrow{TPR_mean} + \thisrow{TPR_std}, 
		col sep=comma
	    ] {means_stds.csv};

	    \addplot [
		blue,
        dashed,
		forget plot
	    ] table [
		x=Threshold,
		y expr=\thisrow{TPR_mean} - \thisrow{TPR_std}, 
		col sep=comma
	    ] {means_stds.csv};

	    \addplot [
		no markers,
		blue,
		mark options={fill=white},
	    ] table [
		x=Threshold,
		y=TPR_mean,
		col sep=comma
	    ] {means_stds.csv};

	    \addplot [
		red,
        dashed,
		forget plot
	    ] table [
		x=Threshold,
		y expr=\thisrow{FPR_mean} + \thisrow{FPR_std}, 
		col sep=comma
	    ] {means_stds.csv};

	    \addplot [
		red,
        dashed,
		forget plot
	    ] table [
		x=Threshold,
		y expr=\thisrow{FPR_mean} - \thisrow{FPR_std}, 
		col sep=comma
	    ] {means_stds.csv};

	    \addplot [
		no markers,
		red,
		mark options={fill=white}
	    ] table [
		x=Threshold,
		y=FPR_mean,
		col sep=comma
	    ] {means_stds.csv};
        \legend{TPR Mean, FPR Mean}

	    \end{axis}
	\end{tikzpicture}
    \end{subfigure}
    \hfill
    \begin{subfigure}[b]{0.2\textwidth}
        \begin{tikzpicture}[scale=0.45]
	    \begin{axis}[
		xlabel={Threshold},
        clip=false,
		ylabel={Rate},
		legend pos=south east,
        xmajorgrids=true,
        ymajorgrids=true,
        grid style=dashed,
	    ]

	    \addplot [
		blue,
        dashed,
		forget plot
	    ] table [
		x=Threshold,
		y expr=\thisrow{TPR_mean} + \thisrow{TPR_std}, 
		col sep=comma
	    ] {means_stdsDIF.csv};

	    \addplot [
		blue,
        dashed,
		forget plot
	    ] table [
		x=Threshold,
		y expr=\thisrow{TPR_mean} - \thisrow{TPR_std}, 
		col sep=comma
	    ] {means_stdsDIF.csv};

	    \addplot [
		no markers,
		blue,
		mark options={fill=white},
	    ] table [
		x=Threshold,
		y=TPR_mean,
		col sep=comma
	    ] {means_stdsDIF.csv};

	    \addplot [
		red,
        dashed,
		forget plot
	    ] table [
		x=Threshold,
		y expr=\thisrow{FPR_mean} + \thisrow{FPR_std}, 
		col sep=comma
	    ] {means_stdsDIF.csv};

	    \addplot [
		red,
        dashed,
		forget plot
	    ] table [
		x=Threshold,
		y expr=\thisrow{FPR_mean} - \thisrow{FPR_std}, 
		col sep=comma
	    ] {means_stdsDIF.csv};

	    \addplot [
		no markers,
		red,
		mark options={fill=white}
	    ] table [
		x=Threshold,
		y=FPR_mean,
		col sep=comma
	    ] {means_stdsDIF.csv};
        \legend{TPR Mean, FPR Mean}

	    \end{axis}
	\end{tikzpicture}
    \end{subfigure}
    
    \begin{subfigure}[b]{0.2\textwidth}
        \begin{tikzpicture}[scale=0.45]
	    \begin{axis}[
            xlabel={Threshold},
            ylabel={Accuracy},
            clip=false,
            xmin=10, xmax=61000,
            ymin=0.5, ymax=0.7,
            legend pos=south west,
            xmajorgrids=true,
            ymajorgrids=true,
            grid style=dashed,
        ]

	    \addplot[blue, mark=*, mark size=1pt] coordinates {
        (2061, 0.5841396795)
        (2211, 0.5892968292)
        (2377, 0.5914309644)
        (2563, 0.5950545233)
        (2774, 0.5956923893)
        (3013, 0.5979757836)
        (3288, 0.5977993562)
        (3608, 0.5977348892)
        (3986, 0.5963777418)
        (4437, 0.594006128)
        (4986, 0.5928695408)
        (5667, 0.5914411338)
        (8413, 0.5854934339)
        (14091, 0.5773437734)
        (27777, 0.5610342645)
        (60386, 0.5410740238)
        };

        \addplot[red, mark=*, mark size=1pt] coordinates {
        (2061, 0.6547181056)
        (2211, 0.6583009821)
        (2377, 0.6647406312)
        (2563, 0.6689952887)
        (2774, 0.6727681664)
        (3013, 0.6767038886)
        (3288, 0.6776301402)
        (3608, 0.6808160314)
        (3986, 0.6828347857)
        (4437, 0.6869808848)
        (4986, 0.6881480263)
        (5667, 0.6891624928)
        (8413, 0.6905908906)
        (14091, 0.6927724847)
        (27777, 0.6930541029)
        (60386, 0.6927521275)
        };
        
	    \legend{Filtered, Noise-injection}

	    \end{axis}
	\end{tikzpicture}
    \caption*{~~~~~~~~~~NN}
    \end{subfigure}
    \hfill
    \begin{subfigure}[b]{0.2\textwidth}
        \begin{tikzpicture}[scale=0.45]
	    \begin{axis}[
            xlabel={Threshold},
            ylabel={Accuracy},
            clip=false,
            xmin=3000, xmax=28000,
            ymin=0.5, ymax=0.7,
            legend pos=south west,
            xmajorgrids=true,
            ymajorgrids=true,
            grid style=dashed,
        ]

	    \addplot[blue, mark=*, mark size=1pt] coordinates {
        (3659, 0.6335465954)
        (3903, 0.6385917939)
        (4179, 0.6435521658)
        (4497, 0.648125754)
        (4867, 0.6539852252)
        (5302, 0.6592000494)
        (5824, 0.6658908083)
        (6460, 0.6714279514)
        (7249, 0.6750617348)
        (8253, 0.6725306373)
        (9547, 0.6674650999)
        (11238, 0.6586979353)
        (13467, 0.6468330805)
        (16408, 0.6272867735)
        (20311, 0.5962522007)
        (26814, 0.5303118573)
        };

        \addplot[red, mark=*, mark size=1pt] coordinates {
        (3659, 0.6471689756)
        (3903, 0.6537918403)
        (4179, 0.6593731)
        (4497, 0.6630679415)
        (4867, 0.6662029395)
        (5302, 0.6697010077)
        (5824, 0.673372099)
        (6460, 0.6776063901)
        (7249, 0.6833606775)
        (8253, 0.6863565858)
        (9547, 0.6887214092)
        (11238, 0.6895594413)
        (13467, 0.6930541167)
        (16408, 0.6940244711)
        (20311, 0.6941330341)
        (26814, 0.6936715933)
        };
        
	    \legend{Filtered, Noise-injection}

	    \end{axis}
	\end{tikzpicture}
    \caption*{~~~~~~~~~~DIF}
    \end{subfigure}

    \caption{Comparison of true-positive and false-positive rate values together with the accuracy score depending on the NN and DIF filter parameter for the CNN model and the N-Caltech101. In both cases, it can be seen that our method performs better against the filters regardless of the parameters.}
    
\label{fig:plots_tpr_fpr_accuracy}
\end{figure}
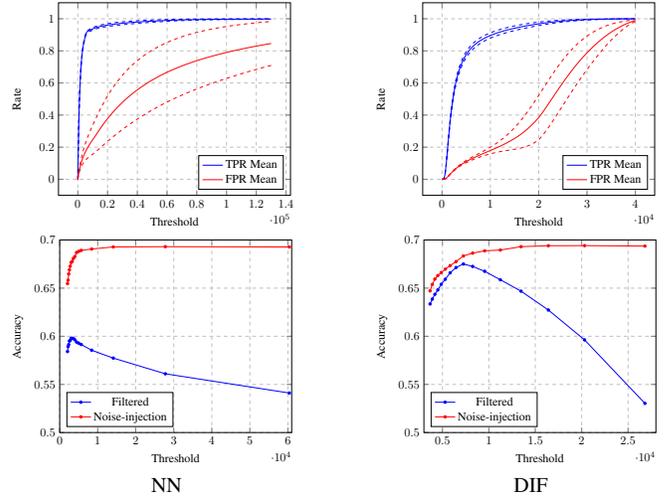

We conducted our experiments in the same way as for classification, training the models on four different training sets (\textit{Original}, \textit{Filtrated}, \textit{Noise=1Hz/px}, and \textit{Noise-injection} – see Figure \ref{fig:detection_plots}). We then evaluated the averaged results across the entire range of noise frequencies using different filters (\textit{NN}, \textit{DIF}, and \textit{EDnCNN} – see Table \ref{table:detection}). As the main metric, we employed the mean average precision (mAP).

From the results shown in Figure \ref{fig:detection_plots}, we can observe that even for the more complex task of detection, our learning method ensures performance stability regardless of the input noise frequency. In the detection setting, the models trained on a single noise frequency (1Hz/px) best results within that particular noise range. Meanwhile, similarly to the classification task, all methods apart from ours gradually degrade as the noise frequency increases.

When evaluating the results with filtering, our method again achieves the best outcomes for both models. The differences amount to 5\% mAP with NN filtering between our approach and the \textit{Filtered} for the CNN model, and 5.5\% mAP for the ViT model. This demonstrates that our \textit{noise-injection} approach can also be successfully applied to other, more complex tasks such as object detection.

\section{Ablation over different filter parameters}

In order to evaluate the impact of filter parameter settings on denoising effectiveness and overall network performance, an ablation study was conducted for both the NN and DIF methods, using the baseline CNN model and the N-Clatech101 dataset. The results are presented in Figure \ref{fig:plots_tpr_fpr_accuracy}.
Due to the fact that the EDnCNN only returns the probability of an event to be noise, which should be set to 50\%, it was excluded from this analysis.

Initially, the True Positive Rate (TPR) and the False Positive Rate (FPR) were evaluated for various filter thresholds. Subsequent to this, representative thresholds were selected, and the average accuracy was assessed on the full test set under different noise intensities.

The TPR and FPR plots provide a clear indication that there is no threshold value at which the filter perfectly removes noise while preserving all true events. The analysis suggests that the optimal filter parameters are 5000 for NN and 7000 for DIF, where the TPR is already relatively high and the FPR remains sufficiently low.  

The accuracy plots as a function of the filter parameter demonstrate that models trained on \textit{Filtered} data achieve their highest accuracy at these optimal thresholds, while performance decreases for extreme parameter values. Notably, regardless of the filtering strategy, models trained with our proposed \textit{Noise-injection} approach consistently outperform those trained on filtered data, even when using the best possible filter parameters. Furthermore, the results indicate that our method performs increasingly well as the filter thresholds rise, i.e. as fewer events are removed. This suggests that aggressive filtering may negatively impact performance by eliminating critical information.  

The findings demonstrate that, even under optimal filtering conditions, the proposed method consistently achieves superior results, thereby highlighting its robustness and ability to effectively handle noisy event data.

\section{Real Noise Analysis}

\input{sec/real_noise_results_plot}

In order to verify that our method also works with real noise, we employed genuine noise obtained from an event camera. To achieve this, an event stream was recorded from a sensor observing a static scene under constant illumination, thereby ensuring that the recorded data were solely related to noise. The noise intensity was varied by adjusting the sensor’s event generation threshold.

For the evaluation, we used the same models as in Section \ref{subsubsec:training}, trained on the Original, Filtered, Noise=1Hz/px and Noise-injection datasets (based on a Poisson process approximation). The results for the N-Caltech101 dataset, tested on all models, are presented in Fig. \ref{fig:real_noise_plots}.

As shown, our method consistently achieves the best and most stable performance across the CNN, ViT, and SNN models. For the GCN model, the results significantly exceed those of the Original and Filtered models, with only the Noise=1Hz/px model achieving higher accuracy under comparable noise levels (around 1Hz/px). For the ViT model, performance at lower noise frequencies is slightly better (by approximately 0.5\%) using the Filtered approach; however, at higher noise levels, our method exhibits a clear and substantial advantage.

These findings demonstrate that our approach is also applicable to real noise from an event camera. Crucially, the performance of models trained on artificially generated disturbances successfully translates to real-world noise.

\section{Timing results}

Filtering, as an additional pre-processing step, requires additional computational resources and increases the overall processing time of the system. Although event reduction through filtering simplifies the generation of representations due to a reduced number of events, in the case of representations such as Event Count Image or Voxel Grid, these operations can be fully parallelised.

We performed an analysis of the average number of events processed for each filtering method, and the results are presented in Table \ref{table:timing}. For NN and DIF filtering, we used a single Intel Xeon Platinum 8268 CPU processor core, while EDnCNN filtering was performed on an NVIDIA A100 GPU and used 16 AMD EPYC 7742 processor cores.

As can be seen from the analysis, the use of the EDnCNN network significantly drops the number of events processed compared to other methods, which has a significant impact on system latency in real-time applications. For the remaining methods, although the number of events processed per second is significantly higher, values below 2 million events per second remain relatively low compared to modern event sensors capable of generating hundreds of millions of events per second. The solution to this problem is hardware acceleration of filters using SoC FPGA systems, as demonstrated with the DIF filter. However, it should be noted that each additional pre-processing element implemented on the FPGA increases the memory and computational requirements. In our approach, the absence of filter implementation on FPGA could contribute positively to resource utilisation in embedded systems.

\begin{table}[htp]
\centering
\resizebox{0.2\textwidth}{!}{%
\begin{tabular}{@{}cc@{}}
\toprule
Filtration & Events per second \\ \midrule
NN         & 1412501.03        \\
DIF        & 1854422.15        \\
EDnCNN     & 8387.66           \\ \bottomrule
\end{tabular}%
}
\caption{Number of events processed per second using NN, DIF and EDnCNN filtering.}
\label{table:timing}
\end{table}

\section{Detail results}
\label{sec:detail_results}

Tables \ref{tab:orig_sup}, \ref{tab:fil_sup}, \ref{tab:1_sup} and \ref{tab:noise_sup} present detailed top-1 and top-3 accuracy results for all versions of the training data: \textit{Original}, \textit{Filtered}, \textit{Noise=1Hz/px} and \textit{Noise-injection}. The average and std values are presented in Table \ref{tab:big_table}.

\begin{figure*}[htp]
    \centering
   	\includegraphics[width=\textwidth]{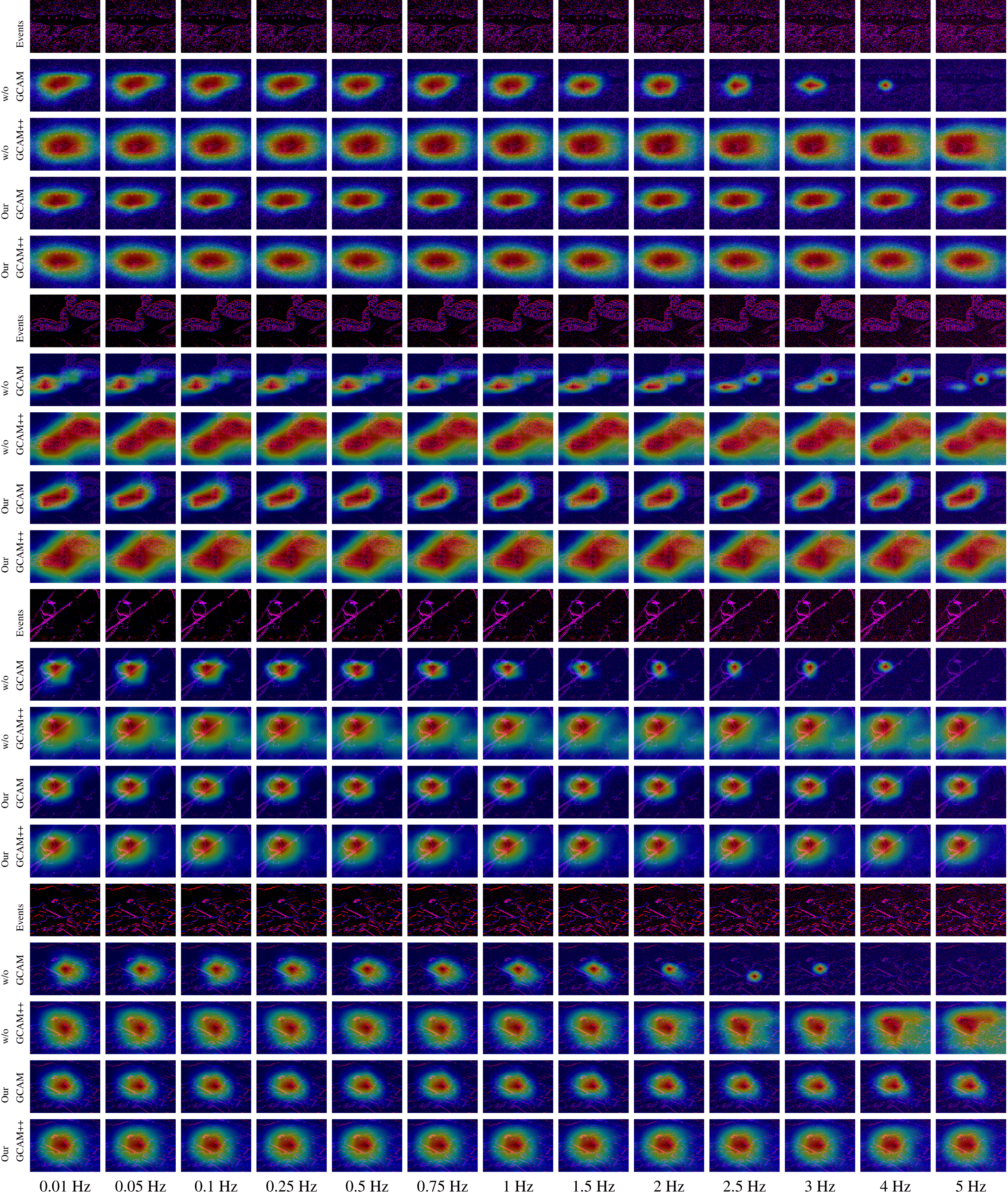}
\caption{Qualitative comparison using GradCAM and GradCam++ between trained CNN model on Mini N-ImageNet with and without our Noise-injection method for different noise levels.}
\label{fig:nimagenet_supp}
\end{figure*}

\begin{figure*}[htp]
    \centering
    \includegraphics[width=\textwidth]{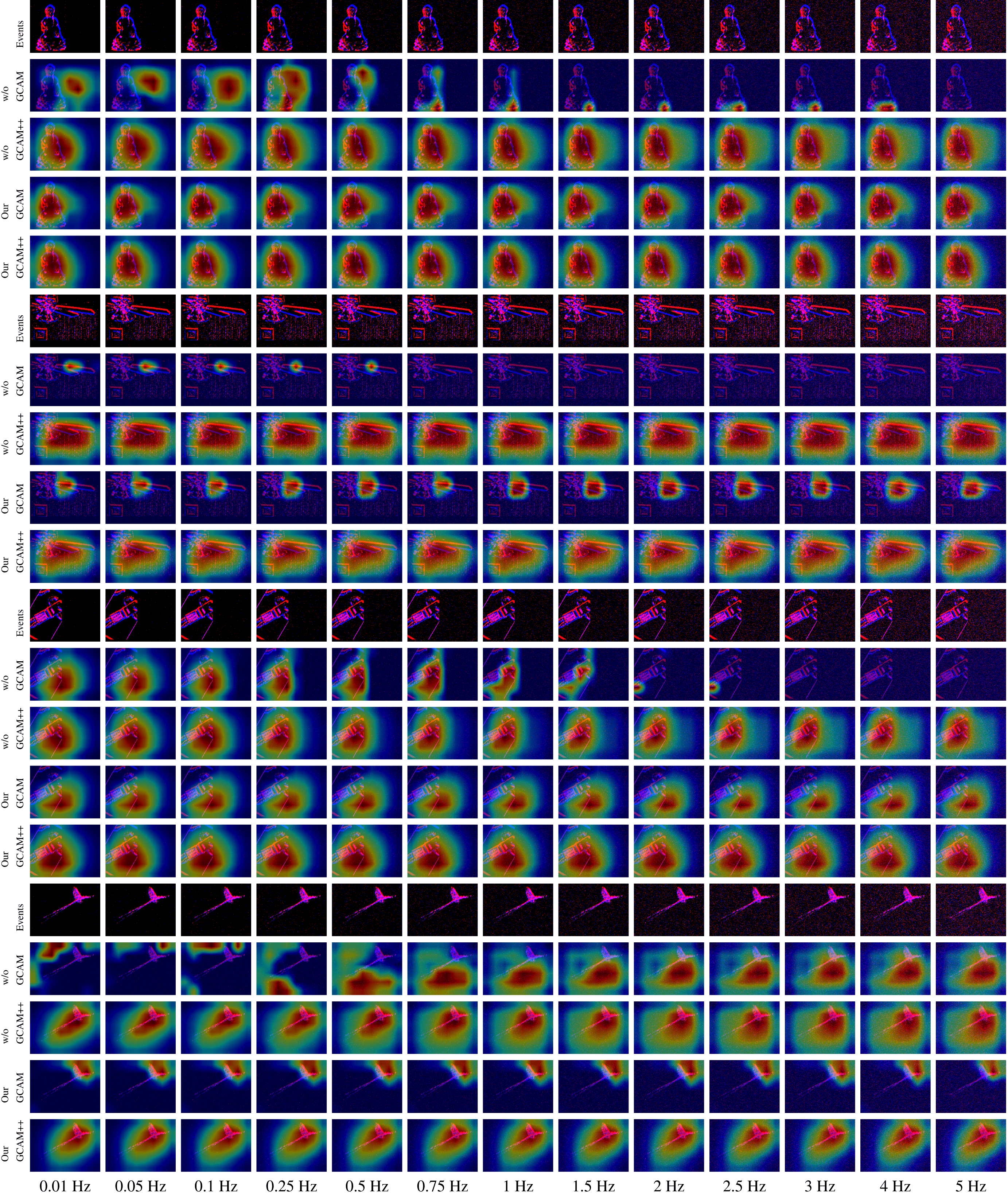}
\caption{Qualitative comparison using GradCAM and GradCam++ between trained CNN model on N-Caltech101 with and without our Noise-injection method for different noise levels.}
\label{fig:ncaltech_supp}
\end{figure*}

Figures \ref{fig:nimagenet_supp} and \ref{fig:ncaltech_supp} illustrate additional GradCAM and GradCAM++ activation map examples for the N-Caltech101 and Mini N-ImageNet datasets, comparing our \textit{Noise-injection} method with no method applied. All examples show consistent patterns: increased input noise results in reduced activation intensity for GradCAM and dispersed activations in background regions for GradCAM++. In contrast, activations remain stable when using our \textit{Noise-injection} method.

\begin{table*}[htp]
\resizebox{\textwidth}{!}{%
\begin{tabular}{lcccccccccccccccccccc}
\cline{2-21}
\multicolumn{1}{l|}{} &
  \multicolumn{8}{c|}{N-Caltech101} &
  \multicolumn{4}{c|}{N-Cars} &
  \multicolumn{8}{c}{N-ImageNet} \\ \cline{2-21} 
\multicolumn{1}{c|}{} &
  \multicolumn{2}{c|}{NN} &
  \multicolumn{2}{c|}{EDnCNN} &
  \multicolumn{2}{c|}{DIF} &
  \multicolumn{2}{c|}{w/o} &
  \multicolumn{1}{c|}{NN} &
  \multicolumn{1}{c|}{EDnCNN} &
  \multicolumn{1}{c|}{DIF} &
  \multicolumn{1}{c|}{w/o} &
  \multicolumn{2}{c|}{NN} &
  \multicolumn{2}{c|}{EDnCNN} &
  \multicolumn{2}{c|}{DIF} &
  \multicolumn{2}{c}{w/o} \\ \cline{1-1}
\multicolumn{1}{l|}{Noise level} &
  Acc. &
  \multicolumn{1}{c|}{Acc. top-3} &
  Acc. &
  \multicolumn{1}{c|}{Acc. top-3} &
  Acc. &
  \multicolumn{1}{c|}{Acc. top-3} &
  Acc. &
  \multicolumn{1}{c|}{Acc. top-3} &
  \multicolumn{1}{c|}{Acc.} &
  \multicolumn{1}{c|}{Acc.} &
  \multicolumn{1}{c|}{Acc.} &
  \multicolumn{1}{c|}{Acc.} &
  Acc. &
  \multicolumn{1}{c|}{Acc. top-3} &
  Acc. &
  \multicolumn{1}{c|}{Acc. top-3} &
  Acc. &
  \multicolumn{1}{c|}{Acc. top-3} &
  Acc. &
  Acc. top-3 \\ \hline
\multicolumn{21}{c}{\textbf{CNN}} \\ \hline
\multicolumn{1}{l|}{0.01} &
  \cellcolor[HTML]{EFEFEF}63.84 &
  76.88 &
  \cellcolor[HTML]{EFEFEF}59.55 &
  74.16 &
  \cellcolor[HTML]{EFEFEF}63.28 &
  76.46 &
  \cellcolor[HTML]{EFEFEF}67.18 &
  \multicolumn{1}{c|}{78.83} &
  \cellcolor[HTML]{EFEFEF}88.71 &
  80.47 &
  \cellcolor[HTML]{EFEFEF}86.41 &
  \multicolumn{1}{c|}{77.89} &
  \cellcolor[HTML]{EFEFEF}36.74 &
  54.99 &
  \cellcolor[HTML]{EFEFEF}14.24 &
  26.01 &
  \cellcolor[HTML]{EFEFEF}33.01 &
  49.94 &
  \cellcolor[HTML]{EFEFEF}42.91 &
  61.14 \\
\hline
\end{tabular}
}

\caption{Detail quantitative results of different models trained on \textit{Original} data verions on the N-Caltech101 \cite{orchard2015converting}, N-Cars \cite{sironi2018hats} and Mini N-ImageNet \cite{Kim_2021_ICCV} datasets, using NN \cite{delbruck2008frame}, EDnCNN \cite{baldwin2020event}, DIF \cite{kowalczyk2023interpolation} and without filtration.}
\label{tab:orig_sup}
\end{table*}

\begin{table*}[htp]
\resizebox{\textwidth}{!}{%
\begin{tabular}{lcccccccccccccccccccc}
\cline{2-21}
\multicolumn{1}{l|}{} &
  \multicolumn{8}{c|}{N-Caltech101} &
  \multicolumn{4}{c|}{N-Cars} &
  \multicolumn{8}{c}{N-ImageNet} \\ \cline{2-21} 
\multicolumn{1}{c|}{} &
  \multicolumn{2}{c|}{NN} &
  \multicolumn{2}{c|}{EDnCNN} &
  \multicolumn{2}{c|}{DIF} &
  \multicolumn{2}{c|}{w/o} &
  \multicolumn{1}{c|}{NN} &
  \multicolumn{1}{c|}{EDnCNN} &
  \multicolumn{1}{c|}{DIF} &
  \multicolumn{1}{c|}{w/o} &
  \multicolumn{2}{c|}{NN} &
  \multicolumn{2}{c|}{EDnCNN} &
  \multicolumn{2}{c|}{DIF} &
  \multicolumn{2}{c}{w/o} \\ \cline{1-1}
\multicolumn{1}{l|}{Noise level} &
  Acc. &
  \multicolumn{1}{c|}{Acc. top-3} &
  Acc. &
  \multicolumn{1}{c|}{Acc. top-3} &
  Acc. &
  \multicolumn{1}{c|}{Acc. top-3} &
  Acc. &
  \multicolumn{1}{c|}{Acc. top-3} &
  \multicolumn{1}{c|}{Acc.} &
  \multicolumn{1}{c|}{Acc.} &
  \multicolumn{1}{c|}{Acc.} &
  \multicolumn{1}{c|}{Acc.} &
  Acc. &
  \multicolumn{1}{c|}{Acc. top-3} &
  Acc. &
  \multicolumn{1}{c|}{Acc. top-3} &
  Acc. &
  \multicolumn{1}{c|}{Acc. top-3} &
  Acc. &
  Acc. top-3 \\ \hline
\multicolumn{21}{c}{\textbf{CNN}} \\ \hline
\multicolumn{1}{l|}{0.01} &
  \cellcolor[HTML]{EFEFEF}68.73 &
  80.99 &
  \cellcolor[HTML]{EFEFEF}67.09 &
  79.79 &
  \cellcolor[HTML]{EFEFEF}68.45 &
  80.24 &
  \cellcolor[HTML]{EFEFEF}68.80 &
  \multicolumn{1}{c|}{81.57} &
  \cellcolor[HTML]{EFEFEF}88.43 &
  80.86 &
  \cellcolor[HTML]{EFEFEF}86.72 &
  \multicolumn{1}{c|}{88.60} &
  \cellcolor[HTML]{EFEFEF}42.97 &
  61.50 &
  \cellcolor[HTML]{EFEFEF}19.70 &
  33.52 &
  \cellcolor[HTML]{EFEFEF}39.97 &
  58.72 &
  \cellcolor[HTML]{EFEFEF}45.83 &
  64.94 \\
\hline
\end{tabular}
}

\caption{Detail quantitative results of different models trained on \textit{Filtered} data verions on the N-Caltech101 \cite{orchard2015converting}, N-Cars \cite{sironi2018hats} and Mini N-ImageNet \cite{Kim_2021_ICCV} datasets, using NN \cite{delbruck2008frame}, EDnCNN \cite{baldwin2020event}, DIF \cite{kowalczyk2023interpolation} and without filtration.}
\label{tab:fil_sup}
\end{table*}

\begin{table*}[htp]
\resizebox{\textwidth}{!}{%
\begin{tabular}{lcccccccccccccccccccc}
\cline{2-21}
\multicolumn{1}{l|}{} &
  \multicolumn{8}{c|}{N-Caltech101} &
  \multicolumn{4}{c|}{N-Cars} &
  \multicolumn{8}{c}{N-ImageNet} \\ \cline{2-21} 
\multicolumn{1}{c|}{} &
  \multicolumn{2}{c|}{NN} &
  \multicolumn{2}{c|}{EDnCNN} &
  \multicolumn{2}{c|}{DIF} &
  \multicolumn{2}{c|}{w/o} &
  \multicolumn{1}{c|}{NN} &
  \multicolumn{1}{c|}{EDnCNN} &
  \multicolumn{1}{c|}{DIF} &
  \multicolumn{1}{c|}{w/o} &
  \multicolumn{2}{c|}{NN} &
  \multicolumn{2}{c|}{EDnCNN} &
  \multicolumn{2}{c|}{DIF} &
  \multicolumn{2}{c}{w/o} \\ \cline{1-1}
\multicolumn{1}{l|}{Noise level} &
  Acc. &
  \multicolumn{1}{c|}{Acc. top-3} &
  Acc. &
  \multicolumn{1}{c|}{Acc. top-3} &
  Acc. &
  \multicolumn{1}{c|}{Acc. top-3} &
  Acc. &
  \multicolumn{1}{c|}{Acc. top-3} &
  \multicolumn{1}{c|}{Acc.} &
  \multicolumn{1}{c|}{Acc.} &
  \multicolumn{1}{c|}{Acc.} &
  \multicolumn{1}{c|}{Acc.} &
  Acc. &
  \multicolumn{1}{c|}{Acc. top-3} &
  Acc. &
  \multicolumn{1}{c|}{Acc. top-3} &
  Acc. &
  \multicolumn{1}{c|}{Acc. top-3} &
  Acc. &
  Acc. top-3 \\ \hline
\multicolumn{21}{c}{\textbf{CNN}} \\ \hline
\multicolumn{1}{l|}{0.01} &
  \cellcolor[HTML]{EFEFEF}65.24 &
  77.97 &
  \cellcolor[HTML]{EFEFEF}63.11 &
  77.55 &
  \cellcolor[HTML]{EFEFEF}64.89 &
  77.87 &
  \cellcolor[HTML]{EFEFEF}66.84 &
  \multicolumn{1}{c|}{79.59} &
  \cellcolor[HTML]{EFEFEF}86.31 &
  76.46 &
  \cellcolor[HTML]{EFEFEF}85.47 &
  \multicolumn{1}{c|}{89.64} &
  \cellcolor[HTML]{EFEFEF}35.04 &
  51.74 &
  \cellcolor[HTML]{EFEFEF}15.76 &
  26.92 &
  \cellcolor[HTML]{EFEFEF}32.77 &
  49.70 &
  \cellcolor[HTML]{EFEFEF}40.20 &
  58.29 \\
\hline
\end{tabular}
}

\caption{Detail quantitative results of different models trained on \textit{Noise=1Hz/px} data verions on the N-Caltech101 \cite{orchard2015converting}, N-Cars \cite{sironi2018hats} and Mini N-ImageNet \cite{Kim_2021_ICCV} datasets, using NN \cite{delbruck2008frame}, EDnCNN \cite{baldwin2020event}, DIF \cite{kowalczyk2023interpolation} and without filtration.}
\label{tab:1_sup}
\end{table*}

\begin{table*}[htp]
\resizebox{\textwidth}{!}{%
\begin{tabular}{lcccccccccccccccccccc}
\cline{2-21}
\multicolumn{1}{l|}{} &
  \multicolumn{8}{c|}{N-Caltech101} &
  \multicolumn{4}{c|}{N-Cars} &
  \multicolumn{8}{c}{N-ImageNet} \\ \cline{2-21} 
\multicolumn{1}{c|}{} &
  \multicolumn{2}{c|}{NN} &
  \multicolumn{2}{c|}{EDnCNN} &
  \multicolumn{2}{c|}{DIF} &
  \multicolumn{2}{c|}{w/o} &
  \multicolumn{1}{c|}{NN} &
  \multicolumn{1}{c|}{EDnCNN} &
  \multicolumn{1}{c|}{DIF} &
  \multicolumn{1}{c|}{w/o} &
  \multicolumn{2}{c|}{NN} &
  \multicolumn{2}{c|}{EDnCNN} &
  \multicolumn{2}{c|}{DIF} &
  \multicolumn{2}{c}{w/o} \\ \cline{1-1}
\multicolumn{1}{l|}{Noise level} &
  Acc. &
  \multicolumn{1}{c|}{Acc. top-3} &
  Acc. &
  \multicolumn{1}{c|}{Acc. top-3} &
  Acc. &
  \multicolumn{1}{c|}{Acc. top-3} &
  Acc. &
  \multicolumn{1}{c|}{Acc. top-3} &
  \multicolumn{1}{c|}{Acc.} &
  \multicolumn{1}{c|}{Acc.} &
  \multicolumn{1}{c|}{Acc.} &
  \multicolumn{1}{c|}{Acc.} &
  Acc. &
  \multicolumn{1}{c|}{Acc. top-3} &
  Acc. &
  \multicolumn{1}{c|}{Acc. top-3} &
  Acc. &
  \multicolumn{1}{c|}{Acc. top-3} &
  Acc. &
  Acc. top-3 \\ \hline
\multicolumn{21}{c}{\textbf{CNN}} \\ \hline
\multicolumn{1}{l|}{0.01} &
  \cellcolor[HTML]{EFEFEF}68.83 &
  80.48 &
  \cellcolor[HTML]{EFEFEF}67.88 &
  80.31 &
  \cellcolor[HTML]{EFEFEF}69.14 &
  80.31 &
  \cellcolor[HTML]{EFEFEF}69.54 &
  \multicolumn{1}{c|}{80.65} &
  \cellcolor[HTML]{EFEFEF}89.01 &
  82.12 &
  \cellcolor[HTML]{EFEFEF}87.77 &
  \multicolumn{1}{c|}{91.50} &
  \cellcolor[HTML]{EFEFEF}42.83 &
  61.46 &
  \cellcolor[HTML]{EFEFEF}21.54 &
  34.73 &
  \cellcolor[HTML]{EFEFEF}42.19 &
  60.48 &
  \cellcolor[HTML]{EFEFEF}45.65 &
  64.54 \\
\hline
\end{tabular}
}

\caption{Detail quantitative results of different models trained on \textit{Noise-injection} data verions on the N-Caltech101 \cite{orchard2015converting}, N-Cars \cite{sironi2018hats} and Mini N-ImageNet \cite{Kim_2021_ICCV} datasets, using NN \cite{delbruck2008frame}, EDnCNN \cite{baldwin2020event}, DIF \cite{kowalczyk2023interpolation} and without filtration.}

\label{tab:noise_sup}
\end{table*}


\end{document}